\documentclass{article}

\usepackage[preprint]{neurips_2026}
\usepackage[ruled]{algorithm2e}
\usepackage{amsfonts}       % blackboard math symbols
\usepackage{amsmath}
\usepackage{amssymb}
\usepackage{booktabs}       % professional-quality tables
\usepackage[T1]{fontenc}    % use 8-bit T1 fonts
\usepackage[pdftex]{graphicx}
\usepackage{hyperref}       % hyperlinks
\usepackage[utf8]{inputenc} % allow utf-8 input
\usepackage{microtype}      % microtypography
\usepackage{multirow}
\usepackage{nicefrac}       % compact symbols for 1/2, etc.
\usepackage{placeins}
\usepackage{subcaption}
\usepackage{url}            % simple URL typesetting
\usepackage{xcolor}         % colors
\usepackage[capitalize]{cleveref}

\newcommand{\se}[1]{{\scriptsize$\,\pm\,$#1}}
\crefname{algocf}{Algorithm}{Algorithms}

\title{Quantized Keys Steal Attention: Bias Correction for KV-Cache Compression in Video Diffusion}

\author{
  Tuna Tuncer\textsuperscript{1,2} \quad
  Felix Becker\textsuperscript{2,$\dagger$\:\:\:\:\:\:}
  Thomas Pfeil\textsuperscript{2,$\dagger$} \\
  \textsuperscript{1}Technical University of Munich \\
  \textsuperscript{2}Tensordyne \\
  \texttt{tuna.tuncer@tum.de} \quad
  \texttt{felix.becker@tensordyne.ai} \quad
  \texttt{thomas.pfeil@tensordyne.ai}
}

\begin{document}

\maketitle

\makeatletter
\if@preprint
  \begingroup
  \renewcommand\thefootnote{}
  \footnotetext{$^\dagger$ Felix Becker and Thomas Pfeil jointly supervised this work.}
  \endgroup
\fi
\makeatother

% =============================================================================
% Abstract
% =============================================================================
\begin{abstract}
Chunk-wise autoregressive video diffusion models rely on a KV cache of previously generated chunks to avoid redundant computation, but this cache quickly becomes a memory bottleneck as videos grow longer.
Methods that quantize the KV cache to low bitwidths reduce memory pressure but
degrade video quality. We show that a key driver of this
degradation is a systematic bias in attention weights: due to the convexity of the exponential in softmax
attention, quantization noise inflates the contribution of cached keys, a phenomenon we call the \emph{Jensen bias}.
This effect causes quantized keys to steal attention mass from
the unquantized current chunk.
We derive a per-attention-score correction that removes this bias in expectation, computed on the fly from the quantization step sizes of the cached keys and the query norm.
Using a second-order Taylor approximation, the additional computational overhead is negligible, and no additional memory is
needed alongside the cache.
Evaluated on MAGI-1, SkyReels-V2, and HY-WorldPlay at INT2 quantization, our
correction recovers most of the quality lost to aggressive quantization, reaching near-BF16 video quality, and can outperform INT4 quantization while using 50\% less memory.
\end{abstract}

% =============================================================================
% Section 1: Introduction
% =============================================================================
\section{Introduction}
\label{sec:introduction}

Video diffusion models have made remarkable progress in
generating short, high-fidelity
clips~\citep{yang2025cogvideox, kong2025hunyuanvideo, wan2025}.
Recent work on video generation models has introduced chunk-wise autoregressive video diffusion,
where each chunk of frames is denoised independently and attends to previously generated chunks~\citep{chen2024diffusion, yin2025causvid, magi2025, chen2025skyreelsv2infinitelengthfilmgenerative, sun2025worldplay}. To avoid recomputing the key and value representations
of past chunks at every denoising step, autoregressive models store them in a KV
cache and reuse them across subsequent chunks. In this setting, the KV cache acts as the model's temporal memory:
it determines how much previously generated visual context remains
available when simulating the next chunk of a video or world trajectory.

\begin{figure}[t]
  \centering
  \includegraphics[width=\textwidth]{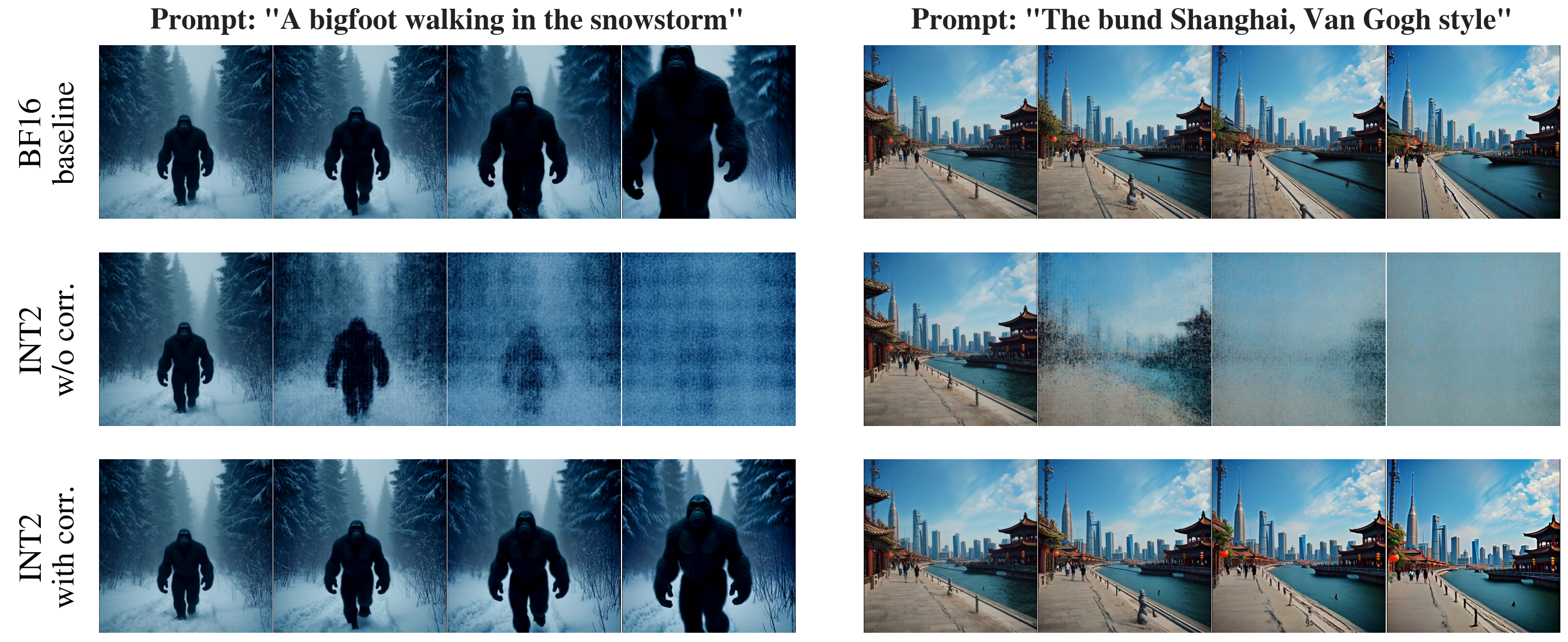}
\caption{Qualitative comparison on MAGI-1 for two representative prompts. Columns show successive frames from the same generated video. From top to bottom: BF16 baseline; asymmetric INT2 (QuaRot+RTN) KV-cache quantization of both keys and values; same quantized setting with our correction. INT2 quantization quickly destroys subject and scene structure, whereas our correction substantially recovers the BF16-like visual quality and temporal consistency.}
\label{fig:qualitative_comparison}
\end{figure}

\begin{figure}[t]
  \centering
  \includegraphics[width=\textwidth]{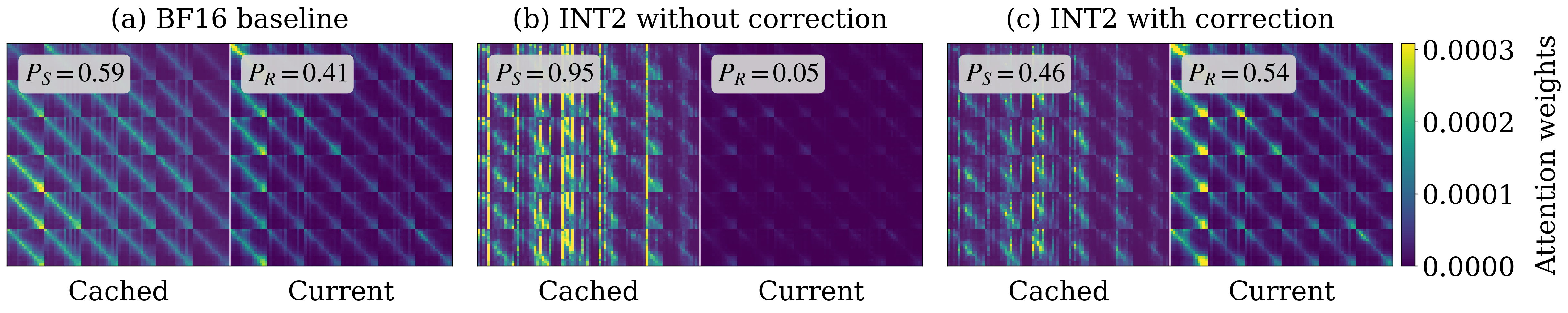}
  \caption{Attention weights for MAGI-1 for the
    prompt ``a person'' under INT2 KV-cache quantization. The visualization is taken from a representative layer, time step, and attention head.
    Panel \textbf{(b)} shows that relative to the BF16 baseline in \textbf{(a)}, quantization increases attention weights in the cached block of tokens and decreases them in the current chunk. This effect is quantified by the \emph{attention masses} $P_\mathcal{S}$ and $P_\mathcal{R}$ of the cached token blocks and current chunks. \textbf{(c)} shows that our correction largely restores the original attention weights.
    }
  \label{fig:motivation_distribution}
\end{figure}

\begin{figure}[t]
  \centering
  \includegraphics[width=\textwidth]{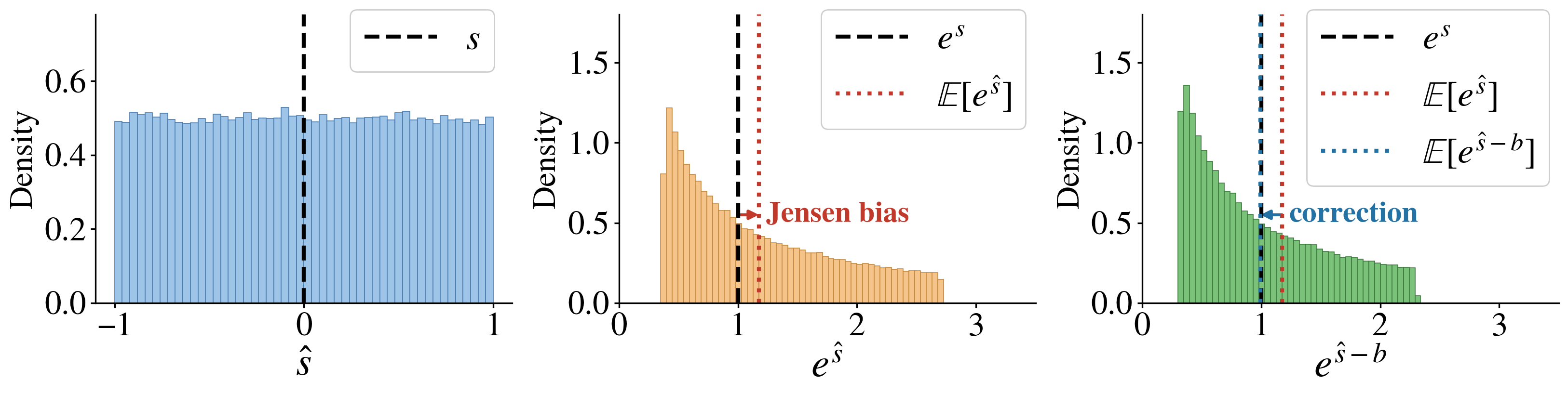}
  \caption{Illustration of the \emph{Jensen bias} and its correction on a single attention score.
  \textbf{Left:} Quantization noise $\delta \sim \mathrm{Uniform}[-\Delta/2,\Delta/2]$ with zero mean produces a noisy score $\hat{s} = s + \delta$ centered at $s$.
  \textbf{Center:} After exponentiation the distribution becomes right-skewed: its mean $\mathbb{E}[e^{\hat{s}}]$ strictly exceeds $e^{s}$ by the so-called \emph{Jensen bias}.
  \textbf{Right:} Subtracting a correction $b$ shifts the mean $\mathbb{E}[e^{\hat{s}-b}]$ closer to $e^{s}$, largely removing the systematic Jensen bias.}
  \label{fig:jensen_bias}
\end{figure}

To further reduce the attention cost, MAGI-1~\citep{magi2025} attends to a sliding window of the last \textit{n} cached chunks, yielding linear instead of quadratic scaling in video length. This design introduces a fundamental memory–context trade-off: increasing the window size improves temporal consistency by providing more past context, but also increases the size of the KV cache proportionally.
Due to memory capacity, memory bandwidth, and latency constraints in practical systems, the window size must be limited, restricting the temporal information available to the model and degrading long-range consistency~\citep{xi2026quantvideogenautoregressivelong,
samuel2026fastautoregressivevideodiffusion}.

KV-cache quantization directly targets the underlying memory bottleneck by compressing the cached keys and values to lower bitwidths, thereby relaxing this trade-off: the same memory budget can support a larger context window, or a fixed window can be stored more efficiently. Prior
work on KV-cache quantization for LLM inference~\citep{liu2024kivi, hooper2024kvquant, ashkboos2024quarot}
has established effective techniques down to 2-bit precision. For autoregressive video models,
we find that INT4 KV-cache quantization preserves reasonable quality, whereas reducing to INT2 leads
to severely distorted frames
% (\cref{fig:qualitative_comparison,fig:qualitative_comparison_skyreels,fig:qualitative_comparison_hywp}).
(\cref{fig:qualitative_comparison}, \cref{fig:qualitative_comparison_skyreels}, and
\cref{fig:qualitative_comparison_hywp}).

We identify a shift of \emph{attention mass} toward cached tokens under aggressive quantization as an important source of this degradation (see example in \cref{fig:motivation_distribution} and definition in \cref{sec:setup}). This shift is consistent across layers, heads, denoising steps, and prompts, and correlates with poor video quality (\cref{fig:qualitative_comparison}).
Integer quantization introduces approximately zero-mean noise into the cached keys, leaving pre-softmax attention scores unbiased in expectation. However, the exponential in softmax breaks this symmetry: due to its convexity, positive deviations are amplified more than equally large negative deviations are suppressed. As a result, a symmetric score-level noise distribution becomes right-skewed after exponentiation, with its mean systematically exceeding the exponential of the original unquantized score (\cref{fig:jensen_bias}). We refer to this systematic, convexity-induced inflation as the \emph{Jensen bias}, as it is an instance of the Jensen gap studied in probability theory~\citep{gao2020bounds}. In chunk-wise autoregressive video diffusion, this bias inflates the cached-token contribution to the softmax partition sum at the expense of the current chunk.

Our correction directly targets the Jensen bias. Because the bias is systematic,
it can be estimated from quantities available at inference time and subtracted
from the cached-key attention scores before the softmax. This restores the
balance between cached and current tokens without retraining or modifying the
quantized KV cache values (\cref{fig:motivation_distribution}).

Our contributions are as follows:
\begin{itemize}
  \item We identify the Jensen bias, a systematic inflation induced by KV-cache
  quantization, in which zero-mean cached-key score perturbations
  inflate the expected cached-token softmax contribution and shift
  attention mass away from the unquantized current chunk.

  \item We derive a theoretically grounded per-attention-score correction
  and show that a simple second-order Taylor approximation yields an
  effective, practical formula with negligible overhead.

  \item We demonstrate consistent benchmark improvements across multiple
  models and quantization schemes, validating the proposed correction
  from attention-level diagnostics through to end-to-end video quality.
\end{itemize}

% =============================================================================
% Section 2: Related Work
% =============================================================================
\section{Related Work}
\label{sec:related_work}

\paragraph{KV-cache quantization for LLMs.}
The KV cache is a well-known memory bottleneck in long-context
LLM inference \cite{kwon2023vllm}, and a
growing body of work addresses it through quantization:
KIVI~\citep{liu2024kivi} provides an early systematic study of
KV cache element distributions, observing that keys exhibit
channel-wise outliers while values do not, and exploits this
asymmetry to achieve tuning-free 2-bit KV quantization.
KVQuant~\citep{hooper2024kvquant} combines per-channel key
quantization with non-uniform datatypes calibrated to the
empirical KV distribution and explicit isolation of outlier
entries, pushing KV caches below 4 bits with minimal
perplexity loss. QuaRot~\citep{ashkboos2024quarot} applies Hadamard rotations to
spread channel-wise outliers before quantization, enabling
outlier-free 4-bit inference.
TurboQuant~\citep{zandieh2025turboquant} similarly leverages random rotations, framing KV-cache compression as an online vector quantization problem and applying scalar quantization in the rotated space to achieve near-optimal distortion at low bitwidth.
AsymKV~\citep{asymkv2024} observes that model loss is more
sensitive to key quantization than value quantization and
proposes layer-wise asymmetric bit allocation, supporting our
focus on key cache quantization.
Our work is orthogonal to the approaches above in that we do not improve the quantization scheme itself, but instead
analytically correct the systematic bias in the attention weights introduced by any such scheme.

\paragraph{Attention sensitivity and correction.}
Several works have studied how quantization and other
perturbations affect the attention mechanism.
\citet{pandey2023softmaxbias} show that
quantizing the softmax computation introduces a large bias in
the softmax output, degrading accuracy in generative models,
and propose an offline correction that can be folded into the
quantization parameters.
Our work targets a different source of bias, focusing on KV-cache
quantization rather than softmax quantization.
KVLinC~\citep{saxena2025kvlinckvcache} is conceptually closest to our
approach: it introduces trainable linear correction adapters to
compensate errors from quantized keys.
In contrast, our correction is training-free and analytically
derived.
SageAttention~\citep{zhang2024sageattention2} smooths queries by
subtracting channel means and adds a correction term to the
scores. However, this targets quantization-friendliness of the
$QK^\top$ product rather than the systematic bias from
exponentiation.
\citet{yao2024timestepawarecorrectionquantizeddiffusion} propose time step-aware
corrections for quantized diffusion models, demonstrating that
structure-aware corrections can substantially reduce quantization
degradation, a principle our per-attention-score correction shares.

\paragraph{Autoregressive video diffusion and efficient caching.}
Chunk-wise autoregressive video diffusion models generate videos
by denoising successive chunks that attend to previously
generated chunks through a KV
cache~\citep{chen2024diffusion, yin2025causvid, magi2025, chen2025skyreelsv2infinitelengthfilmgenerative, sun2025worldplay}.
Because the cache grows with each new chunk, a growing body of
work aims to reduce its cost through cache compression and
eviction~\citep{ma2026flowcachingautoregressivevideo,
chen2026contextforcingconsistentautoregressive,
samuel2026fastautoregressivevideodiffusion},
sparse
attention~\citep{lv2026lightforcing},
or direct quantization of the cached
states~\citep{xi2026quantvideogenautoregressivelong}.
Among these, QuantVideoGen~\citep{xi2026quantvideogenautoregressivelong}
is most directly related to our approach: it applies training-free KV-cache
quantization using semantic-aware smoothing and progressive
residual quantization to reduce the quantization error itself.
Our approach is complementary: rather than reducing the quantization error, we
analytically correct the bias it introduces in softmax attention.
We validate this complementarity empirically in \cref{tab:main_results}, where composing the two methods on MAGI-1 yields the best overall results.

% =============================================================================
% Section 3: Preliminaries
% =============================================================================
\section{Preliminaries}
\label{sec:preliminaries}

\paragraph{Integer quantization.}
Integer quantization maps a floating-point value to a discrete
grid defined by a \emph{scale} $\Delta$, also known as the step size between
adjacent grid levels, and a \emph{zero-point} $z$.
Given a $B$-bit quantization target, each element $x$ is mapped
to
\begin{equation}
  x_q = \mathrm{clamp}\:\!\bigl(
    \lfloor x / \Delta \rceil + z,\; 0,\; 2^B{-}1
  \bigr),
  \label{eq:quantize}
\end{equation}
where $\lfloor \cdot \rceil$ denotes rounding to nearest (RTN), and is reconstructed as $\hat{x} = (x_q - z) \cdot \Delta$.
The round-trip $x \mapsto x_q \mapsto \hat{x}$ introduces an
additive error $\epsilon = \hat{x} - x$ that is bounded by
$|\epsilon| \leq \Delta/2$.
In practice, both $\Delta$ and $z$ are
chosen to cover the full $[\min, \max]$ range of the value
being quantized.

\paragraph{Quantization granularity.}
The scale and zero-point can be shared at different
granularities.
In per-tensor quantization, one $(\Delta, z)$ pair is shared across
an entire tensor.
In per-token quantization, each token has its own
$(\Delta_i, z_i)$.
Group-wise per-token quantization further divides each
token's $d$ channels into groups of size $g$, with an
independent $(\Delta_{i,j}, z_{i,j})$ per group $j$.
The smaller the group of values sharing $(\Delta, z)$, the smaller the quantization error, but the larger the overall memory footprint.

\paragraph{Hadamard rotation.}
\label{sec:bg_quarot}
Key vectors in transformer models often exhibit channel-wise
outliers, i.e.\ a few channels have much larger magnitudes than the
rest~\citep{dettmers2022llmint8,ashkboos2024quarot}.
These outliers inflate the quantization step size $\Delta$,
degrading precision for all other channels.
QuaRot~\citep{ashkboos2024quarot} spreads the outlier energy across all
channels by applying a
randomized Hadamard rotation $H \in \mathbb{R}^{d \times d}$
(with $H^\top H = I$) to both keys and queries. The resulting distribution is more uniform, allowing for lower quantization errors.
Because $H$ is orthogonal, the attention scores are preserved:
$(Hq)^\top(Hk) = q^\top k$.
For all ablation studies, we use such a Hadamard rotation before quantization, since this results in overall best quantized video quality.

\paragraph{Token Structure and Attention Decomposition.}
In autoregressive video diffusion, each chunk of video frames is encoded into a latent representation and patchified into a grid of spatio-temporal tokens before entering the transformer. Depending on the model and resolution, this results in several thousand tokens per chunk.
At each denoising step, every query in the current chunk attends
to two groups of keys: (i)~the keys of the current chunk, which
are computed in full precision at every step, and (ii)~the keys
of previously generated chunks, which were written to a
KV cache once each chunk finished denoising and are
reused without recomputation.
The attention score matrix therefore decomposes into two blocks:
a \emph{current} block of tokens (current-chunk queries $\times$
current-chunk keys) and a \emph{cached} block of tokens (current-chunk
queries $\times$ cached keys).

We now turn to the effect of quantization on this attention mechanism and derive a correction that compensates for the resulting bias in the softmax computation.

% =============================================================================
% Section 4: Method
% =============================================================================
\section{Method}
\label{sec:method}

We analyze the effect of KV-cache quantization on softmax attention and
show that it introduces a systematic bias that inflates the
contribution of cached keys. Based on this analysis, we derive a correction term that removes this bias in expectation, and present
a practical approximation suitable for efficient implementation.

\subsection{Quantization Bias in Softmax Attention}
% \subsection{Score-Space Noise Under Quantization}
\label{sec:setup}

Consider a single attention head with dimension~$d$.
For a query vector $q \in \mathbb{R}^d$ and key vectors
$k_i \in \mathbb{R}^d$, where $i$ is the token index, the attention score and attention weight for token $i$ are
\begin{equation}
  s_i = \frac{q^\top k_i}{\sqrt{d}}, \qquad
  p_i = \frac{e^{s_i}}{\sum_{j=1}^{N} e^{s_j}}.
  \label{eq:attention}
\end{equation}

Recall from \cref{sec:preliminaries} that in autoregressive video
generation, tokens from previously generated chunks are quantized and stored in
the KV cache, while tokens of the current chunk have not yet been quantized.
Let $\mathcal{S}$ denote the set of quantized \emph{cached} key
indices and $\mathcal{R}$ the set of unquantized \emph{current-chunk} key indices,
so that $\{1,\dots,N\} = \mathcal{S}\cup \mathcal{R}$.
We define the partition sums
\begin{equation}
  Z_{\mathcal{S}} = \sum_{i \in \mathcal{S}} e^{s_i}, \qquad
  Z_{\mathcal{R}} = \sum_{i \in \mathcal{R}} e^{s_i}, \qquad
  Z = Z_{\mathcal{S}} + Z_{\mathcal{R}}.
  \label{eq:partition}
\end{equation}
We also define the total attention mass on the
cached block,
\begin{equation}
  P_{\mathcal{S}}
  = \sum_{i \in \mathcal{S}} p_i
  = \frac{Z_{\mathcal{S}}}{Z_{\mathcal{S}} + Z_{\mathcal{R}}},
  \label{eq:prob_mass}
\end{equation}
which measures how much attention mass is assigned to
cached keys, and is what we ultimately care about when reasoning
about attention stealing.
For a representative example of attention stealing, compare left to middle panel in \cref{fig:motivation_distribution}.

\paragraph{Quantization noise model.}
Let $\Delta_{i,c}$ denote the quantization step size for
channel~$c$ of cached token~$i$.
The quantize--dequantize round-trip yields
$\hat{k}_i = k_i + \epsilon_i$ for $i \in \mathcal{S}$.
For the per-element error of integer quantization
$\epsilon_i \in \mathbb{R}^d$, we
assume that the components are independent across
channels $c \in \{1,\dots,d\}$ and uniformly distributed~\citep{widrow1996statistical}:
\begin{equation}
  \epsilon_{i,c}
  \sim \mathcal{U}\!\left(
    -\frac{\Delta_{i,c}}{2},\;
    +\frac{\Delta_{i,c}}{2}
  \right).
  \label{eq:noise_model}
\end{equation}

Note that this noise model depends only on the round-to-nearest
quantization operation itself, not on any preprocessing applied to
the keys before quantization (such as Hadamard rotations in QuaRot;
see \cref{app:quarot}).

The quantized attention score is then
\begin{equation}
  \hat{s}_i
  = \frac{q^\top \hat{k}_i}{\sqrt{d}}
  = s_i + \delta_i, \qquad
  \delta_i = \frac{q^\top \epsilon_i}{\sqrt{d}},
  \label{eq:noisy_score}
\end{equation}
where $\delta_i$ is the attention-score noise for key~$i$.
Under the uniform noise model, $\delta_i$ has zero mean and,
by channel independence, its variance is
\begin{equation}
  \sigma_i^2
  = \operatorname{Var}(\delta_i)
  = \frac{1}{12\,d}
    \sum_{c=1}^{d} q_c^2 \, \Delta_{i,c}^2.
  \label{eq:noise_var}
\end{equation}
For unquantized keys $i \in \mathcal{R}$, we have $\hat{s}_i = s_i$.

\label{sec:bias}

% The score-space noise $\delta_i$ introduces a systematic bias in the softmax partition sum, which we now describe.

\paragraph{Jensen bias and attention stealing.}
Consider the quantized cached partition sum
$\hat{Z}_{\mathcal{S}}
= \sum_{i \in \mathcal{S}} e^{s_i + \delta_i}$.
By linearity of expectation:
\begin{equation}
  \mathbb{E}\bigl[\hat{Z}_{\mathcal{S}}\bigr]
  = \sum_{i \in \mathcal{S}}
    e^{s_i} \cdot \mathbb{E}\bigl[e^{\delta_i}\bigr].
  \label{eq:expected_Zhat}
\end{equation}
For each term, Jensen's inequality applied to the convex
function $\exp(\cdot)$ gives
$\mathbb{E}[e^{\delta_i}] \geq e^{\mathbb{E}[\delta_i]} = 1$,
so that
$\mathbb{E}[\hat{Z}_{\mathcal{S}}] \geq Z_{\mathcal{S}}$.
We call this systematic inflation of $\hat{Z}_{\mathcal{S}}$ caused by $\delta_i$ the
\emph{Jensen bias}.
See \cref{fig:jensen_bias} for an illustration of this bias and its correction on a single attention score value.

Since $Z_{\mathcal R}$ is unaffected by key quantization, inflation
of $\hat Z_{\mathcal S}$ can shift attention mass toward cached keys.
We quantify this \emph{attention stealing} as
\begin{equation}
  \Delta P_{\mathcal S}
  =
  \hat P_{\mathcal S} - P_{\mathcal S},
  \qquad
  \hat P_{\mathcal S}
  =
  \frac{\hat Z_{\mathcal S}}
       {\hat Z_{\mathcal S}+Z_{\mathcal R}} .
  \label{eq:attn_stealing}
\end{equation}
Positive values indicate excess attention on the cached block, as
observed in \cref{sec:analysis}.

\subsection{Correction of the Jensen Bias}
\label{sec:computing_correction}

We derive a per-attention-score correction~$b_i$ that counteracts the Jensen bias, applied
only to cached scores ($i \in \mathcal{S}$) and leaving
current-chunk scores $s_i$ ($i \in \mathcal{R}$) unchanged.
As shown in \cref{sec:bias}, each cached token's contribution to the partition sum is individually biased upward:
$\mathbb{E}[e^{s_i+\delta_i}] = e^{s_i}\,\mathbb{E}[e^{\delta_i}] \geq e^{s_i}$.
We correct each token individually by requiring its expected contribution to match the unquantized value:
\begin{equation}
  e^{s_i - b_i} \cdot \mathbb{E}\bigl[e^{\delta_i}\bigr]
  \overset{!}{=}
  e^{s_i}
  \quad \Longrightarrow \quad
  \boxed{\;
    b_i = \log \mathbb{E}\bigl[e^{\delta_i}\bigr].
  \;}
  \label{eq:correction}
\end{equation}
Since every term is individually unbiased, the corrected cached partition sum is unbiased by linearity of expectation:
\begin{equation}
  \mathbb{E}\bigl[\tilde{Z}_{\mathcal{S}}\bigr]
  = \sum_{i \in \mathcal{S}}
    e^{s_i - b_i} \cdot \mathbb{E}\bigl[e^{\delta_i}\bigr]
  = \sum_{i \in \mathcal{S}} e^{s_i}
  = Z_{\mathcal{S}}.
  \label{eq:obj1}
\end{equation}
At inference time, we apply this correction by subtracting $b_i$
from each cached attention score $s_i$ prior to the softmax, leaving scores from the current (unquantized) keys unchanged.
Note that $b_i \geq 0$ always (since
$\mathbb{E}[e^{\delta_i}] \geq 1$ by Jensen's inequality).
Furthermore, $b_i$ increases with the score-space noise, i.e. with $\Delta_{i,c}$.

Since the noise components $\epsilon_{i,c}$ are independent across channels
the expectation $\mathbb{E}[e^{\delta_i}]$ factorizes across dimensions, leading to the exact correction term (for the full derivation, see \cref{app:exact_derivation}):
\begin{equation}
  \boxed{%
    \rule{0pt}{3.4ex}%
    b_i = \sum_{c=1}^{d} \log\!\left(
    \frac{
      \sinh\!\left(\dfrac{q_c\,\Delta_{i,c}}{2\sqrt{d}}\right)
    }{
      \dfrac{q_c\,\Delta_{i,c}}{2\sqrt{d}}
    }
    \right)
    \rule[-1.2ex]{0pt}{0pt}%
    \;
  }
  \label{eq:exact_correction}
\end{equation}
Setting $\alpha_c = q_c \Delta_{i,c} / (2\sqrt{d})$ and using the second-order Taylor expansion $\log(\sinh(\alpha_c)/\alpha_c) \approx \alpha_c^2/6$ for small $|\alpha_c|$, this simplifies to:
\begin{equation}
  \boxed{%
    \rule{0pt}{3.4ex}%
    b_i \approx \frac{1}{24\,d}\sum_{c=1}^{d} q_c^2 \,\Delta_{i,c}^2
    \rule[-1.2ex]{0pt}{0pt}%
    \;
  }
  \label{eq:taylor_correction}
\end{equation}
The Taylor approximation is simple, interpretable, and numerically stable. It shows that the bias scales with both the squared query magnitude and the squared quantization step size. We use this approximation in all experiments.
For a representative example of this proposed correction, compare middle to right panel in \cref{fig:motivation_distribution}.

\paragraph{Connection to the noise variance.}
Comparing \cref{eq:taylor_correction} with \cref{eq:noise_var},
the Taylor correction is exactly half the score-space noise variance:
$b_i \approx \sigma_i^2 / 2$.
This follows from the cumulant generating function (CGF).
For any random variable $X$ with cumulants
$\kappa_1, \kappa_2, \kappa_3, \ldots$, the CGF satisfies
\begin{equation}
  \log \mathbb{E}[e^{X}]
  = \kappa_1
  + \frac{\kappa_2}{2}
  + \frac{\kappa_3}{6}
  + \cdots\,.
  \label{eq:cgf}
\end{equation}
For zero-mean noise ($\kappa_1 = 0$), the leading term is
$\kappa_2 / 2 = \sigma^2 / 2$, which depends only on the
variance and not on the specific noise distribution.
The exact closed-form correction in
\cref{eq:exact_correction} relies on the
uniform noise model of integer quantization, but the
second-order Taylor approximation requires only the
score-space noise variance~$\sigma_i^2$.
This means that extending the correction to other quantization
formats reduces to estimating $\sigma_i^2$ under the
appropriate error model: for floating-point formats such as FP, MXFP, and NVFP,
whose rounding error is proportional to the magnitude of the
quantized value but can be described by approximate additive
noise models~\citep{widrow1996statistical}, one substitutes the
corresponding score-space variance into
$b_i \approx \sigma_i^2 / 2$.

\paragraph{Specialization to grouped per-token quantization.}
In our experimental setting, each token's $d$ channels are
divided into $G = d/g$ groups of size~$g$, and all channels within
group~$j$ share the same step size $\Delta_{i,j}$.
Grouping channels with shared step sizes, and writing
$\|q_j\|^2 = \sum_{c \in \text{group } j} q_c^2$ for the
per-group squared query norm, we obtain
\begin{equation}
  b_i \approx \frac{1}{24d}
  \sum_{j=1}^{G} \Delta_{i,j}^2 \, \|q_j\|^2.
  \label{eq:grouped_correction}
\end{equation}
The same correction extends to QuaRot by replacing $q$ with the
rotated query $Hq$, so that $\|q_j\|^2$ becomes $\|(Hq)_j\|^2$ in
\cref{eq:grouped_correction} (for full derivation, see \cref{app:quarot}).

\subsection{Effective bitwidth and computational complexity}
\label{sec:impl_overhead}

For group-wise quantization with group size $g$, the effective bitwidth is
$B_{\mathrm{eff}} = B + \frac{24}{g}$, accounting for per-group scale stored in
FP8 and zero-point stored in BF16 metadata.
Our correction introduces no additional storage, as it depends only on existing quantization parameters.

The Taylor correction adds an $O(QK \cdot d/g)$ term to attention computation, compared to the standard $O(QK \cdot d)$ cost of $QK^\top$.
Thus, the additional work is smaller by a factor of $g$ and is negligible in practice. In our FlexAttention-based implementation~\citep{dong2024flexattention}
on MAGI-1 with QuaRot+RTN and group size $g{=}32$, the correction adds approximately $5\%$ end-to-end latency overhead relative to the quantized baseline.
For more details about these storage and computation costs, see \cref{app:detailed_cost_breakdown}.

\begin{table}[t]
  \centering
  \caption{
  Effect of the proposed correction for MAGI-1,
  SkyReels-V2, and HY-WorldPlay.
  The correction consistently improves fidelity (PSNR, SSIM, LPIPS)
  and perceptual quality (VBench), recovering much of the degradation
  introduced by quantization.
  RTN and QuaRot+RTN rows use an effective bitwidth of $2.75$ at INT2;
  QVG rows on MAGI-1 use the default QVG configuration, which yields
  an effective bitwidth of approximately $2.52$.
  Standard errors for all metrics are reported in
  \cref{tab:fidelity_se,tab:vbench_aggregate}.
  }
  \label{tab:main_results}
  \small
  \renewcommand{\arraystretch}{0.92}
  \begin{tabular}{llcccccc}
    \toprule
    \textbf{Model}
      & \shortstack{\textbf{Quantization}\\\textbf{scheme}}
      & \textbf{Precision}
      & \shortstack{\textbf{With}\\\textbf{correction}}
      & \textbf{PSNR}$\uparrow$
      & \textbf{SSIM}$\uparrow$
      & \textbf{LPIPS}$\downarrow$
      & \textbf{VBench}$\uparrow$ \\
    \midrule
    \multirow{9}{*}{MAGI-1}
      & ---
      & BF16
      &
      & -- & -- & -- & 78.27 \\
    \cmidrule(lr){2-8}
      & \multirow{2}{*}{RTN}
      & \multirow{2}{*}{INT2}
      & $\times$     & 23.17 & 0.799 & 0.195 & 76.67 \\
      & & & $\checkmark$ & 24.08 & 0.828 & 0.131 & 77.86 \\
    \cmidrule(lr){2-8}
      & \multirow{2}{*}{QuaRot+RTN}
      & \multirow{2}{*}{INT2}
      & $\times$     & 17.10 & 0.630 & 0.453 & 70.24 \\
      & & & $\checkmark$ & 22.97 & 0.801 & 0.165 & 78.02 \\
    \cmidrule(lr){2-8}
      & \multirow{2}{*}{QVG}
      & \multirow{2}{*}{INT2}
      & $\times$     & 23.01 & 0.826 & 0.132 & 77.81 \\
      & & & $\checkmark$ & \textbf{25.29} & \textbf{0.856} & \textbf{0.107} & \textbf{78.23} \\
    \midrule
    \multirow{5}{*}{SkyReels-V2}
      & ---
      & BF16
      &
      & -- & -- & -- & 78.89 \\
    \cmidrule(lr){2-8}
      & \multirow{2}{*}{RTN}
      & \multirow{2}{*}{INT2}
      & $\times$     & 18.38 & 0.670 & 0.383 & 68.83 \\
      & & & $\checkmark$ & 18.39 & 0.743 & 0.269 & \textbf{80.04} \\
    \cmidrule(lr){2-8}
      & \multirow{2}{*}{QuaRot+RTN}
      & \multirow{2}{*}{INT2}
      & $\times$     & 19.20 & 0.708 & 0.319 & 71.44 \\
      & & & $\checkmark$ & \textbf{20.42} & \textbf{0.784} & \textbf{0.202} & 78.58 \\
    \midrule
    \multirow{4}{*}{HY-WorldPlay}
      & \multirow{2}{*}{RTN}
      & \multirow{2}{*}{INT2}
      & $\times$     & 16.96 & 0.573 & 0.390 & -- \\
      & & & $\checkmark$ & 17.15 & 0.577 & 0.311 & -- \\
    \cmidrule(lr){2-8}
      & \multirow{2}{*}{QuaRot+RTN}
      & \multirow{2}{*}{INT2}
      & $\times$     & 17.16 & 0.575 & 0.376 & -- \\
      & & & $\checkmark$ & \textbf{18.27} & \textbf{0.616} & \textbf{0.273} & -- \\
    \bottomrule
  \end{tabular}
\end{table}

\newpage

% =============================================================================
% Section 5: Experiments
% =============================================================================
\section{Experiments}
\label{sec:experiments}
We evaluate the effectiveness of our proposed correction by measuring its impact both on attention behavior and
on end-to-end video quality across multiple metrics and models.

\subsection{Experimental Setup}
\label{sec:setup_exp}

\paragraph{Models.}
We evaluate our method on three autoregressive video diffusion models: MAGI-1~\citep{magi2025} (4.5B), SkyReels-V2~\citep{chen2025skyreelsv2infinitelengthfilmgenerative} (1.3B), and HY-WorldPlay~\citep{sun2025worldplay} (8B). All use chunk-wise generation with KV caching over previously generated chunks. MAGI-1 uses 16 denoising steps with a sliding window annealed from 5 to 2 chunks, SkyReels-V2 uses 50 steps with a 5-chunk window, and HY-WorldPlay uses 4 steps. Unless otherwise noted, all other generation hyperparameters remain at default values.

\paragraph{Quantization configuration.}
We adopt group-wise per-token asymmetric INT2 quantization of key and value states as the default KV-cache compression setting throughout the paper. Unless otherwise noted, we use group size $g = 32$, FP8 E4M3 scales, BF16 zero-points. We evaluate two quantization schemes:
QuaRot+RTN~\citep{ashkboos2024quarot} and plain RTN without
rotation. Additionally, on MAGI-1 we evaluate QuantVideoGen (QVG)~\citep{xi2026quantvideogenautoregressivelong} using its default configuration ($S{=}1$, $B{=}64$, $K{=}256$) to demonstrate that our correction composes with upstream video-aware cache compression.
We apply the Taylor-approximated bias correction from \cref{sec:computing_correction} to all quantization schemes. Unquantized BF16 results serve as the reference outputs for fidelity metrics.

\paragraph{Metrics.}
We report fidelity metrics (PSNR,
SSIM~\citep{wang2004ssim}, and LPIPS~\citep{zhang2018lpips})
to measure the similarity between quantized and BF16 outputs
on identical inputs. We further evaluate generated videos using
the VBench evaluation framework~\citep{huang2023vbench} in the
VBench-Long setting from VBench++~\citep{huang2024vbenchpp}, which adapts the benchmark to
long-form videos. \cref{tab:main_results} reports the aggregate
VBench score; per-dimension results and Quality/Semantic sub-scores
are provided in \cref{app:vbench_dimensions}.

\paragraph{Evaluation data.}
For MAGI-1 and SkyReels-V2, we evaluate on the first 30\% of
prompts from each VBench-Long dimension, generating 10-second
videos (240 frames) and 7-second videos (177 frames),
respectively. We do not evaluate on the full prompt set, as
this is computationally prohibitive across all models and
quantization configurations.
For HY-WorldPlay, we generate 10-second videos (253 frames)
from the 10 image--prompt pairs released in the official
repository~\citep{sun2025worldplay}. We do not report VBench
scores for this model, as its required inputs (image, text
prompt, and per-frame keyboard actions) are not provided by
any VBench suite.

\subsection{Main Results}
\label{sec:main_results}

KV-cache quantization substantially degrades video quality
(\cref{tab:main_results},
\cref{fig:qualitative_comparison}, \cref{fig:qualitative_comparison_skyreels,fig:qualitative_comparison_hywp}).
Our correction improves fidelity metrics (PSNR, SSIM, LPIPS)
and VBench scores across all three models and both
quantization schemes (\cref{tab:main_results}).
Notably, the correction improves every reported metric in
every evaluated configuration, without any model-specific
tuning. On MAGI-1, composing our correction with QVG achieves the best results across all metrics, confirming that the two methods are complementary: QVG reduces the quantization error while our correction removes the residual Jensen bias.

On MAGI-1 and SkyReels-V2, our correction closes the quality gap between INT2 KV-cache quantization and the BF16 baseline (\cref{tab:main_results}). The MAGI-1 per-dimension breakdown in \cref{app:vbench_dimensions} shows that these gains are broad-based across the VBench dimensions. On HY-WorldPlay, where VBench is not applicable, the correction consistently improves fidelity metrics (\cref{tab:main_results}). \Cref{sec:analysis} links these end-to-end gains to attention-level improvements, including reduced quantization-induced attention shift toward cached tokens.

\begin{figure}[t]
  \centering
  \includegraphics[width=0.85\textwidth]{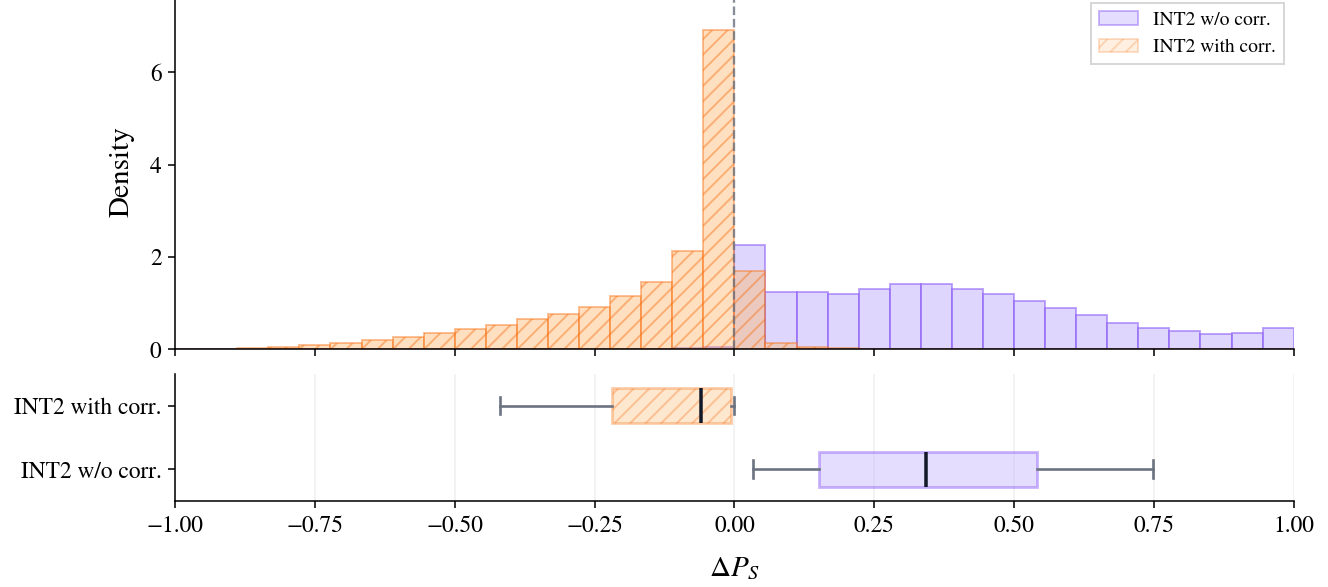}
  \caption{
    Shift in attention mass assigned to the cached block of tokens before (purple) and after (orange) our correction
    on MAGI-1 under INT2 QuaRot+RTN.
    Positive values indicate that the quantized cached tokens steal attention from the current unquantized chunk.
    The median bias is large under INT2 quantization, and our correction significantly reduces this bias toward zero.
    }
  \label{fig:partition_sum}
\end{figure}

\subsection{Ablation studies}
\label{sec:analysis}

We validate our correction by showing that reducing the Jensen bias improves metrics throughout the attention pipeline: attention mass balance, attention weights (JSD; \cref{app:jsd_distributions}), attention outputs (MSE; \cref{app:attn_mse}), and end-to-end video quality (PSNR, VBench).
Together, these evaluations link the score-level Jensen bias to quality degradation and support attention stealing as a key mechanism behind the gains in \cref{sec:main_results}.
All results in this section use MAGI-1 with QuaRot+RTN quantization, the best VBench setting for this model, and are averaged across heads, layers, and denoising steps.

\paragraph{Attention mass shift.}
Attention stealing caused by the Jensen bias is illustrated
in \cref{fig:motivation_distribution}. We quantify this
effect by measuring the shift in attention mass assigned to
cached tokens, $\Delta P_{\mathcal{S}} =
\hat{P}_{\mathcal{S}} - P_{\mathcal{S}}$, aggregated across
all layers, denoising steps, and attention heads.
\Cref{fig:partition_sum} shows that under INT2 quantization,
$\Delta P_{\mathcal{S}}$ is strongly positive, confirming
that cached tokens steal attention mass. Our correction
shifts the distribution back toward zero, though it slightly
over-corrects into negative values, consistent with the
Taylor approximation's behavior at aggressive bitwidths
(\cref{app:exact_derivation}). Corresponding INT4 results
are provided in \cref{app:prob_mass_shift}.

\begin{figure}[t]
  \centering
  \includegraphics[width=0.7\textwidth]{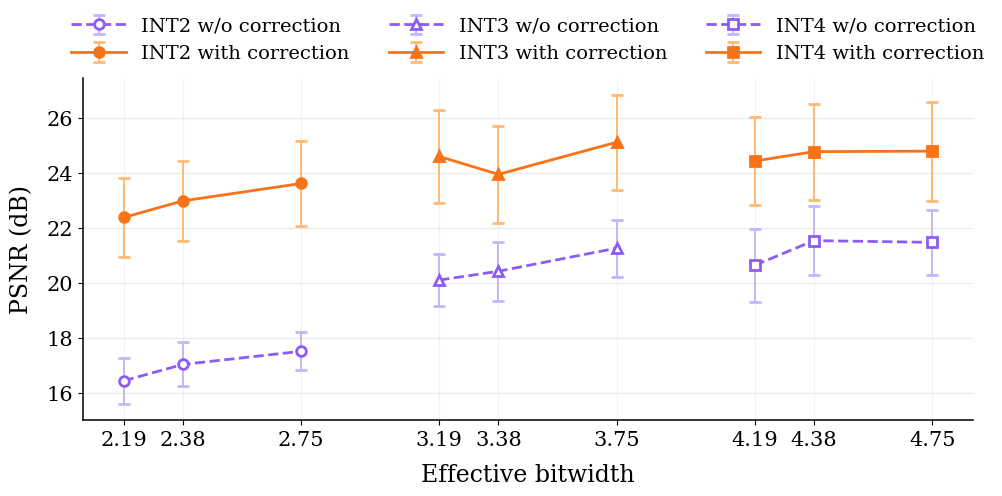}
  \caption{Trade-off between image quality, measured by PSNR, and memory footprint of the KV cache, measured as effective bitwidth per element, on MAGI-1 under quantization. Bitwidths correspond to group sizes $g=\{128,64,32\}$. Whiskers indicate standard error.}
  \label{fig:effective_bitwidth_vs_quality}
\end{figure}

\paragraph{Storage--quality trade-off.}
Our method improves PSNR across all tested group sizes, including the most storage-efficient settings (\cref{fig:effective_bitwidth_vs_quality}). The same trend
holds for SSIM and LPIPS (\cref{app:storage_quality_ssim_lpips}).
Thus, it preserves the group-size-controlled storage--quality trade-off while uniformly shifting it toward higher quality.

Beyond quality gains, our approach also substantially reduces storage and bandwidth requirements at comparable visual fidelity. For example, using 2.19 effective bits with our method outperforms 4.38 effective bits without correction, corresponding to a 50\% reduction in memory cost.

\subsection{Cross-domain experiment: LLM partial prefill}

Although our main experiments target chunk-wise video diffusion, chunked LLM
prefill has a similar cached/current attention structure: a quantized cached
prefix and a multi-token current prefill block appear in the same softmax. We
therefore run a small-scale diagnostic study on three decoder-only LLMs using
LongBench-Pro English prompts~\citep{chen2026longbenchprorealisticcomprehensive}.
We compare BF16, INT2 KV-cache quantization, and INT2 with our Taylor correction
under teacher-forced negative log-likelihood (NLL), using paired model/chunk-size/prompt-length
configurations.

Across the LLM experiments, INT2 generally increases NLL relative to BF16, while the Taylor correction reduces NLL relative to plain INT2. This is consistent with the mechanism studied in our video experiments, but we do not interpret it as a comprehensive LLM benchmark. Details and prompt-length breakdowns are provided in Appendix~\ref{app:llm_partial_prefill}.

% =============================================================================
% Section 6: Discussion and Conclusion
% =============================================================================
\section{Discussion and Conclusion}
\label{sec:conclusion}

We identify a systematic Jensen bias in softmax attention induced by
KV-cache quantization: zero-mean key noise is amplified by the exponential,
inflating cached partition mass and shifting attention away from the
unquantized current chunk. We derive a per-attention-score correction that removes this
bias in expectation and use a second-order Taylor approximation whose cost is
negligible relative to the $QK^\top$ computation. Across MAGI-1, SkyReels-V2,
and HY-WorldPlay, the correction consistently improves fidelity
(PSNR, SSIM, LPIPS) and yields large VBench gains on MAGI-1 and SkyReels-V2,
especially under INT2 quantization.

\paragraph{Limitations \& future work.}
Our experiments focus on chunked autoregressive video diffusion, where a
multi-token current chunk attends to a quantized cached context. This
cached/current structure is central to the attention-mass shift studied here.
Preliminary LLM results suggest that a similar bias can arise in quantized KV
caches. Chunked prefill (where each
prefill contains many current tokens) with KV-cache quantization~\citep{gokhale2025kvpareto}
is therefore a natural target for further exploration. Standard single-token decoding offers less headroom for the correction because many cached tokens compete with only one unquantized current token.

Our correction is unbiased only in expectation and relies on the assumed
zero-mean, approximately uniform quantization-noise model. It works best when
cached attention is spread over enough tokens for score perturbations to
average out. When attention is concentrated on a few cached tokens, the
effective sample size is small and individual noise realizations can dominate,
limiting the correction's gain. Quantizers with nonuniform or biased error may
likewise require a modified derivation.

Because the correction acts only on attention scores, it is orthogonal to the
upstream compression method. Extending it to floating-point formats
such as FP, MXFP, and NVFP, whose non-uniform grids produce a different noise
distribution, remains an open direction.

\begin{ack}
This work was carried out as part of the first author's Master's
thesis at the Technical University of Munich in collaboration with
Tensordyne. We thank Dr.-Ing.\ Victor M.\ van Santen for his
advice and guidance throughout this project, and Prof.\ Dr.-Ing.\
Hussam Amrouch for his supervision at TUM. We further thank
Michael Truong Le and Thomas Elsken at Tensordyne for their
helpful discussions during the course of this work.
\end{ack}

\bibliographystyle{plainnat}
\bibliography{references}

@inproceedings{
liu2024kivi,
title={{KIVI}: A Tuning-Free Asymmetric 2bit Quantization for {KV} Cache},
author={Zirui Liu and Jiayi Yuan and Hongye Jin and Shaochen Zhong and Zhaozhuo Xu and Vladimir Braverman and Beidi Chen and Xia Hu},
booktitle={Forty-first International Conference on Machine Learning},
year={2024},
url={https://openreview.net/forum?id=L057s2Rq8O}
}

@article{hooper2024kvquant,
  title={KVQuant: Towards 10 Million Context Length LLM Inference with KV Cache Quantization},
  author={Hooper, Coleman and Kim, Sehoon and Mohammadzadeh, Hiva and Mahoney, Michael W and Shao, Yakun Sophia and Keutzer, Kurt and Gholami, Amir},
  journal={arXiv preprint arXiv:2401.18079},
  year={2024}
}

@article{wang2004ssim,
  title={Image quality assessment: From error visibility to structural similarity},
  author={Wang, Zhou and Bovik, Alan C. and Sheikh, Hamid R. and Simoncelli, Eero P.},
  journal={IEEE Transactions on Image Processing},
  volume={13},
  number={4},
  pages={600--612},
  year={2004}
}

@article{ashkboos2024quarot,
  title={QuaRot: Outlier-Free 4-Bit Inference in Rotated LLMs},
  author={Ashkboos, Saleh and Mohtashami, Amirkeivan and Croci, Maximilian L and Li, Bo and Jaggi, Martin and Alistarh, Dan and Hoefler, Torsten and Hensman, James},
  journal={arXiv preprint arXiv:2404.00456},
  year={2024}
}

@article{kwon2023vllm,
  title={Efficient Memory Management for Large Language Model Serving with PagedAttention},
  author={Kwon, Woosuk and others},
  journal={arXiv preprint arXiv:2309.06180},
  year={2023}
}

@misc{asymkv2024,
      title={AsymKV: Enabling 1-Bit Quantization of KV Cache with Layer-Wise Asymmetric Quantization Configurations}, 
      author={Qian Tao and Wenyuan Yu and Jingren Zhou},
      year={2024},
      eprint={2410.13212},
      archivePrefix={arXiv},
      primaryClass={cs.LG},
      url={https://arxiv.org/abs/2410.13212}, 
}

@misc{pandey2023softmaxbias,
      title={Softmax Bias Correction for Quantized Generative Models}, 
      author={Nilesh Prasad Pandey and Marios Fournarakis and Chirag Patel and Markus Nagel},
      year={2023},
      eprint={2309.01729},
      archivePrefix={arXiv},
      primaryClass={cs.LG},
      url={https://arxiv.org/abs/2309.01729}, 
}

@misc{saxena2025kvlinckvcache,
      title={KVLinC : KV Cache Quantization with Hadamard Rotation and Linear Correction}, 
      author={Utkarsh Saxena and Kaushik Roy},
      year={2025},
      eprint={2510.05373},
      archivePrefix={arXiv},
      primaryClass={cs.LG},
      url={https://arxiv.org/abs/2510.05373}, 
}

@inproceedings{zhang2024sageattention2,
  title={Sageattention2: Efficient attention with thorough outlier smoothing and per-thread int4 quantization},
  author={Zhang, Jintao and Huang, Haofeng and Zhang, Pengle and Wei, Jia and Zhu, Jun and Chen, Jianfei},
  booktitle={International Conference on Machine Learning (ICML)},
  year={2025}
}

@misc{yao2024timestepawarecorrectionquantizeddiffusion,
      title={Timestep-Aware Correction for Quantized Diffusion Models}, 
      author={Yuzhe Yao and Feng Tian and Jun Chen and Haonan Lin and Guang Dai and Yong Liu and Jingdong Wang},
      year={2024},
      eprint={2407.03917},
      archivePrefix={arXiv},
      primaryClass={cs.CV},
      url={https://arxiv.org/abs/2407.03917}, 
}

@misc{magi2025,
      title={MAGI-1: Autoregressive Video Generation at Scale},
      author={{Sand.ai} and Hansi Teng and Hongyu Jia and Lei Sun and Lingzhi Li and Maolin Li and Mingqiu Tang and Shuai Han and Tianning Zhang and W. Q. Zhang and Weifeng Luo and Xiaoyang Kang and Yuchen Sun and Yue Cao and Yunpeng Huang and Yutong Lin and Yuxin Fang and Zewei Tao and Zheng Zhang and Zhongshu Wang and Zixun Liu and Dai Shi and Guoli Su and Hanwen Sun and Hong Pan and Jie Wang and Jiexin Sheng and Min Cui and Min Hu and Ming Yan and Shucheng Yin and Siran Zhang and Tingting Liu and Xianping Yin and Xiaoyu Yang and Xin Song and Xuan Hu and Yankai Zhang and Yuqiao Li},
      year={2025},
      eprint={2505.13211},
      archivePrefix={arXiv},
      primaryClass={cs.CV},
      url={https://arxiv.org/abs/2505.13211},
}

@misc{sun2025worldplay,
      title={WorldPlay: Towards Long-Term Geometric Consistency for Real-Time Interactive World Modeling}, 
      author={Wenqiang Sun and Haiyu Zhang and Haoyuan Wang and Junta Wu and Zehan Wang and Zhenwei Wang and Yunhong Wang and Jun Zhang and Tengfei Wang and Chunchao Guo},
      year={2025},
      eprint={2512.14614},
      archivePrefix={arXiv},
      primaryClass={cs.CV},
      url={https://arxiv.org/abs/2512.14614}, 
}

@misc{ma2026flowcachingautoregressivevideo,
      title={Flow caching for autoregressive video generation}, 
      author={Yuexiao Ma and Xuzhe Zheng and Jing Xu and Xiwei Xu and Feng Ling and Xiawu Zheng and Huafeng Kuang and Huixia Li and Xing Wang and Xuefeng Xiao and Fei Chao and Rongrong Ji},
      year={2026},
      eprint={2602.10825},
      archivePrefix={arXiv},
      primaryClass={cs.CV},
      url={https://arxiv.org/abs/2602.10825}, 
}

@misc{chen2026contextforcingconsistentautoregressive,
      title={Context Forcing: Consistent Autoregressive Video Generation with Long Context}, 
      author={Shuo Chen and Cong Wei and Sun Sun and Ping Nie and Kai Zhou and Ge Zhang and Ming-Hsuan Yang and Wenhu Chen},
      year={2026},
      eprint={2602.06028},
      archivePrefix={arXiv},
      primaryClass={cs.CV},
      url={https://arxiv.org/abs/2602.06028}, 
}

@misc{samuel2026fastautoregressivevideodiffusion,
      title={Fast Autoregressive Video Diffusion and World Models with Temporal Cache Compression and Sparse Attention}, 
      author={Dvir Samuel and Issar Tzachor and Matan Levy and Micahel Green and Gal Chechik and Rami Ben-Ari},
      year={2026},
      eprint={2602.01801},
      archivePrefix={arXiv},
      primaryClass={cs.CV},
      url={https://arxiv.org/abs/2602.01801}, 
}

@misc{xi2026quantvideogenautoregressivelong,
      title={Quant VideoGen: Auto-Regressive Long Video Generation via 2-Bit KV-Cache Quantization}, 
      author={Haocheng Xi and Shuo Yang and Yilong Zhao and Muyang Li and Han Cai and Xingyang Li and Yujun Lin and Zhuoyang Zhang and Jintao Zhang and Xiuyu Li and Zhiying Xu and Jun Wu and Chenfeng Xu and Ion Stoica and Song Han and Kurt Keutzer},
      year={2026},
      eprint={2602.02958},
      archivePrefix={arXiv},
      primaryClass={cs.LG},
      url={https://arxiv.org/abs/2602.02958}, 
}

@misc{dettmers2022llmint8,
      title={LLM.int8(): 8-bit Matrix Multiplication for Transformers at Scale}, 
      author={Tim Dettmers and Mike Lewis and Younes Belkada and Luke Zettlemoyer},
      year={2022},
      eprint={2208.07339},
      archivePrefix={arXiv},
      primaryClass={cs.LG},
      url={https://arxiv.org/abs/2208.07339}, 
}

@misc{zhang2018lpips,
      title={The Unreasonable Effectiveness of Deep Features as a Perceptual Metric}, 
      author={Richard Zhang and Phillip Isola and Alexei A. Efros and Eli Shechtman and Oliver Wang},
      year={2018},
      eprint={1801.03924},
      archivePrefix={arXiv},
      primaryClass={cs.CV},
      url={https://arxiv.org/abs/1801.03924}, 
}

@misc{huang2023vbench,
      title={VBench: Comprehensive Benchmark Suite for Video Generative Models}, 
      author={Ziqi Huang and Yinan He and Jiashuo Yu and Fan Zhang and Chenyang Si and Yuming Jiang and Yuanhan Zhang and Tianxing Wu and Qingyang Jin and Nattapol Chanpaisit and Yaohui Wang and Xinyuan Chen and Limin Wang and Dahua Lin and Yu Qiao and Ziwei Liu},
      year={2023},
      eprint={2311.17982},
      archivePrefix={arXiv},
      primaryClass={cs.CV},
      url={https://arxiv.org/abs/2311.17982}, 
}

@misc{huang2024vbenchpp,
      title={VBench++: Comprehensive and Versatile Benchmark Suite for Video Generative Models}, 
      author={Ziqi Huang and Fan Zhang and Xiaojie Xu and Yinan He and Jiashuo Yu and Ziyue Dong and Qianli Ma and Nattapol Chanpaisit and Chenyang Si and Yuming Jiang and Yaohui Wang and Xinyuan Chen and Ying-Cong Chen and Limin Wang and Dahua Lin and Yu Qiao and Ziwei Liu},
      year={2024},
      eprint={2411.13503},
      archivePrefix={arXiv},
      primaryClass={cs.CV},
      url={https://arxiv.org/abs/2411.13503}, 
}

@misc{chen2026longbenchprorealisticcomprehensive,
      title={LongBench Pro: A More Realistic and Comprehensive Bilingual Long-Context Evaluation Benchmark}, 
      author={Ziyang Chen and Xing Wu and Junlong Jia and Chaochen Gao and Qi Fu and Debing Zhang and Songlin Hu},
      year={2026},
      eprint={2601.02872},
      archivePrefix={arXiv},
      primaryClass={cs.CL},
      url={https://arxiv.org/abs/2601.02872}, 
}

@article{dubey2024llama3herd,
  title        = {The Llama 3 Herd of Models},
  author       = {Dubey, Abhimanyu and Jauhri, Abhinav and Pandey, Abhinav and Kadian, Abhishek and Al-Dahle, Ahmad and Letman, Aiesha and Mathur, Akhil and Schelten, Alan and Yang, Amy and Fan, Angela and others},
  journal      = {arXiv preprint arXiv:2407.21783},
  year         = {2024},
  url          = {https://arxiv.org/abs/2407.21783}
}

@misc{meta2024llama31modelcard,
  title        = {{Meta Llama 3.1 8B} Model Card},
  author       = {{Meta}},
  year         = {2024},
  howpublished = {\url{https://huggingface.co/meta-llama/Llama-3.1-8B}},
  note         = {Accessed: 2026-05-07}
}

@article{jiang2023mistral7b,
  title        = {Mistral 7B},
  author       = {Jiang, Albert Q. and Sablayrolles, Alexandre and Mensch, Arthur and Bamford, Chris and Chaplot, Devendra Singh and de las Casas, Diego and Bressand, Florian and Lengyel, Gianna and Lample, Guillaume and Saulnier, Lucile and Lavaud, L{\'e}lio Renard and Lachaux, Marie-Anne and Stock, Pierre and Le Scao, Teven and Lavril, Thibaut and Wang, Thomas and Lacroix, Timoth{\'e}e and El Sayed, William},
  journal      = {arXiv preprint arXiv:2310.06825},
  year         = {2023},
  url          = {https://arxiv.org/abs/2310.06825}
}

@misc{mistralai2024mistral7binstructv03,
  title        = {{Mistral-7B-Instruct-v0.3} Model Card},
  author       = {{Mistral AI}},
  year         = {2024},
  howpublished = {\url{https://huggingface.co/mistralai/Mistral-7B-Instruct-v0.3}},
  note         = {Accessed: 2026-05-07}
}

@article{qwen2024qwen25technicalreport,
  title        = {Qwen2.5 Technical Report},
  author       = {{Qwen} and Yang, An and Yang, Baosong and Zhang, Beichen and Hui, Binyuan and Zheng, Bo and Yu, Bowen and Li, Chengyuan and Liu, Dayiheng and Huang, Fei and Wei, Haoran and Lin, Huan and Yang, Jian and Tu, Jianhong and Zhang, Jianwei and Yang, Jianxin and Yang, Jiaxi and Zhou, Jingren and Lin, Junyang and Dang, Kai and others},
  journal      = {arXiv preprint arXiv:2412.15115},
  year         = {2024},
  url          = {https://arxiv.org/abs/2412.15115}
}

@misc{qwen2024qwen2532binstruct,
  title        = {{Qwen2.5-32B-Instruct} Model Card},
  author       = {{Qwen}},
  year         = {2024},
  howpublished = {\url{https://huggingface.co/Qwen/Qwen2.5-32B-Instruct}},
  note         = {Accessed: 2026-05-07}
}

@misc{dong2024flexattention,
      title={Flex Attention: A Programming Model for Generating Optimized Attention Kernels}, 
      author={Juechu Dong and Boyuan Feng and Driss Guessous and Yanbo Liang and Horace He},
      year={2024},
      eprint={2412.05496},
      archivePrefix={arXiv},
      primaryClass={cs.LG},
      url={https://arxiv.org/abs/2412.05496}, 
}

@misc{kong2025hunyuanvideo,
      title={HunyuanVideo: A Systematic Framework For Large Video Generative Models}, 
      author={Weijie Kong and Qi Tian and Zijian Zhang and Rox Min and Zuozhuo Dai and Jin Zhou and Jiangfeng Xiong and Xin Li and Bo Wu and Jianwei Zhang and Kathrina Wu and Qin Lin and Junkun Yuan and Yanxin Long and Aladdin Wang and Andong Wang and Changlin Li and Duojun Huang and Fang Yang and Hao Tan and Hongmei Wang and Jacob Song and Jiawang Bai and Jianbing Wu and Jinbao Xue and Joey Wang and Kai Wang and Mengyang Liu and Pengyu Li and Shuai Li and Weiyan Wang and Wenqing Yu and Xinchi Deng and Yang Li and Yi Chen and Yutao Cui and Yuanbo Peng and Zhentao Yu and Zhiyu He and Zhiyong Xu and Zixiang Zhou and Zunnan Xu and Yangyu Tao and Qinglin Lu and Songtao Liu and Dax Zhou and Hongfa Wang and Yong Yang and Di Wang and Yuhong Liu and Jie Jiang and Caesar Zhong},
      year={2025},
      eprint={2412.03603},
      archivePrefix={arXiv},
      primaryClass={cs.CV},
      url={https://arxiv.org/abs/2412.03603}, 
}

@misc{yang2025cogvideox,
      title={CogVideoX: Text-to-Video Diffusion Models with An Expert Transformer}, 
      author={Zhuoyi Yang and Jiayan Teng and Wendi Zheng and Ming Ding and Shiyu Huang and Jiazheng Xu and Yuanming Yang and Wenyi Hong and Xiaohan Zhang and Guanyu Feng and Da Yin and Yuxuan Zhang and Weihan Wang and Yean Cheng and Bin Xu and Xiaotao Gu and Yuxiao Dong and Jie Tang},
      year={2025},
      eprint={2408.06072},
      archivePrefix={arXiv},
      primaryClass={cs.CV},
      url={https://arxiv.org/abs/2408.06072}, 
}

@misc{wan2025,
      title={Wan: Open and Advanced Large-Scale Video Generative Models}, 
      author={{Team Wan} and Ang Wang and Baole Ai and Bin Wen and Chaojie Mao and Chen-Wei Xie and Di Chen and Feiwu Yu and Haiming Zhao and Jianxiao Yang and Jianyuan Zeng and Jiayu Wang and Jingfeng Zhang and Jingren Zhou and Jinkai Wang and Jixuan Chen and Kai Zhu and Kang Zhao and Keyu Yan and Lianghua Huang and Mengyang Feng and Ningyi Zhang and Pandeng Li and Pingyu Wu and Ruihang Chu and Ruili Feng and Shiwei Zhang and Siyang Sun and Tao Fang and Tianxing Wang and Tianyi Gui and Tingyu Weng and Tong Shen and Wei Lin and Wei Wang and Wei Wang and Wenmeng Zhou and Wente Wang and Wenting Shen and Wenyuan Yu and Xianzhong Shi and Xiaoming Huang and Xin Xu and Yan Kou and Yangyu Lv and Yifei Li and Yijing Liu and Yiming Wang and Yingya Zhang and Yitong Huang and Yong Li and You Wu and Yu Liu and Yulin Pan and Yun Zheng and Yuntao Hong and Yupeng Shi and Yutong Feng and Zeyinzi Jiang and Zhen Han and Zhi-Fan Wu and Ziyu Liu},
      year={2025},
      eprint={2503.20314},
      archivePrefix={arXiv},
      primaryClass={cs.CV},
      url={https://arxiv.org/abs/2503.20314}, 
}

@misc{chen2024diffusion,
      title={Diffusion Forcing: Next-token Prediction Meets Full-Sequence Diffusion}, 
      author={Boyuan Chen and Diego Marti Monso and Yilun Du and Max Simchowitz and Russ Tedrake and Vincent Sitzmann},
      year={2024},
      eprint={2407.01392},
      archivePrefix={arXiv},
      primaryClass={cs.LG},
      url={https://arxiv.org/abs/2407.01392}, 
}

@misc{yin2025causvid,
      title={From Slow Bidirectional to Fast Autoregressive Video Diffusion Models}, 
      author={Tianwei Yin and Qiang Zhang and Richard Zhang and William T. Freeman and Fredo Durand and Eli Shechtman and Xun Huang},
      year={2025},
      eprint={2412.07772},
      archivePrefix={arXiv},
      primaryClass={cs.CV},
      url={https://arxiv.org/abs/2412.07772}, 
}

@misc{lv2026lightforcing,
      title={Light Forcing: Accelerating Autoregressive Video Diffusion via Sparse Attention}, 
      author={Chengtao Lv and Yumeng Shi and Yushi Huang and Ruihao Gong and Shen Ren and Wenya Wang},
      year={2026},
      eprint={2602.04789},
      archivePrefix={arXiv},
      primaryClass={cs.CV},
      url={https://arxiv.org/abs/2602.04789}, 
}

@ARTICLE{widrow1996statistical,
  author={Widrow, B. and Kollar, I. and Ming-Chang Liu},
  journal={IEEE Transactions on Instrumentation and Measurement}, 
  title={Statistical theory of quantization}, 
  year={1996},
  volume={45},
  number={2},
  pages={353-361},
  doi={10.1109/19.492748}}

@misc{zandieh2025turboquant,
      title={TurboQuant: Online Vector Quantization with Near-optimal Distortion Rate}, 
      author={Amir Zandieh and Majid Daliri and Majid Hadian and Vahab Mirrokni},
      year={2025},
      eprint={2504.19874},
      archivePrefix={arXiv},
      primaryClass={cs.LG},
      url={https://arxiv.org/abs/2504.19874}, 
}

@misc{gokhale2025kvpareto,
      title={KV Pareto: Systems-Level Optimization of KV Cache and Model Compression for Long Context Inference}, 
      author={Sai Gokhale and Devleena Das and Rajeev Patwari and Ashish Sirasao and Elliott Delaye},
      year={2025},
      eprint={2512.01953},
      archivePrefix={arXiv},
      primaryClass={cs.LG},
      url={https://arxiv.org/abs/2512.01953}, 
}

@article{gao2020bounds,
  title={Bounds on the Jensen Gap, and Implications for Mean-Concentrated Distributions},
  author={Gao, Xiang and Sitharam, Meera and Roitberg, Adrian E.},
  journal={arXiv preprint arXiv:1712.05267},
  year={2020},
  doi={10.48550/arXiv.1712.05267},
  url={https://arxiv.org/abs/1712.05267}
}

@misc{chen2025skyreelsv2infinitelengthfilmgenerative,
      title={SkyReels-V2: Infinite-length Film Generative Model}, 
      author={Guibin Chen and Dixuan Lin and Jiangping Yang and Chunze Lin and Junchen Zhu and Mingyuan Fan and Hao Zhang and Sheng Chen and Zheng Chen and Chengcheng Ma and Weiming Xiong and Wei Wang and Nuo Pang and Kang Kang and Zhiheng Xu and Yuzhe Jin and Yupeng Liang and Yubing Song and Peng Zhao and Boyuan Xu and Di Qiu and Debang Li and Zhengcong Fei and Yang Li and Yahui Zhou},
      year={2025},
      eprint={2504.13074},
      archivePrefix={arXiv},
      primaryClass={cs.CV},
      url={https://arxiv.org/abs/2504.13074}, 
}

%%%%%%%%%%%%%%%%%%%%%%%%%%%%%%%%%%%%%%%%%%%%%%%%%%%%%%%%%%%%

\clearpage
\appendix

\setcounter{figure}{0}
\renewcommand{\thefigure}{A\arabic{figure}}

\section{Exact Correction: Full Derivation}
\label{app:exact_derivation}
 
We derive the exact formula for
$b_i = \log \mathbb{E}[e^{\delta_i} \mid \{\Delta_{i,c}\}]$
under the uniform quantization noise model of \cref{sec:setup}.
 
Recall that
$\delta_i = \sum_{c=1}^{d} q_c \, \epsilon_{i,c} / \sqrt{d}$,
where the $\epsilon_{i,c}$ are independent with
$\epsilon_{i,c} \sim \mathcal{U}(-\Delta_{i,c}/2,
+\Delta_{i,c}/2)$.
By independence across channels, the moment generating function
factorizes:
\begin{equation}
  \mathbb{E}\bigl[e^{\delta_i}\bigr]
  = \prod_{c=1}^{d}
    \mathbb{E}\!\left[
      \exp\!\left(\frac{q_c\,\epsilon_{i,c}}{\sqrt{d}}\right)
    \right].
  \label{eq:app_mgf_factorize}
\end{equation}
For each channel~$c$, we evaluate the scalar MGF.
Let $t_c = q_c / \sqrt{d}$ for brevity.
Since $\epsilon_{i,c} \sim \mathcal{U}(-\Delta_{i,c}/2,\;
+\Delta_{i,c}/2)$:
\begin{align}
  \mathbb{E}\bigl[e^{t_c \, \epsilon_{i,c}}\bigr]
  &= \frac{1}{\Delta_{i,c}}
     \int_{-\Delta_{i,c}/2}^{+\Delta_{i,c}/2}
     e^{t_c \, u} \, du
  \nonumber \\
  &= \frac{\sinh(t_c \, \Delta_{i,c} / 2)}
          {t_c \, \Delta_{i,c} / 2}.
  \label{eq:app_scalar_mgf}
\end{align}
Taking the product over all channels and then the logarithm
yields the exact correction:
\begin{equation}
  b_i
  = \sum_{c=1}^{d} \log\!\left(
    \frac{
      \sinh\!\left(\dfrac{q_c\,\Delta_{i,c}}{2\sqrt{d}}\right)
    }{
      \dfrac{q_c\,\Delta_{i,c}}{2\sqrt{d}}
    }
  \right).
  \label{eq:app_exact_correction}
\end{equation}
 
A naive implementation of this formula is numerically unstable
($\sinh$ overflows for large arguments) and computationally
expensive ($O(d)$ operations per score entry, matching the
attention score computation itself).
We therefore seek a cheaper approximation.
 
\paragraph{Taylor approximation.}
Let $\alpha_c = q_c \, \Delta_{i,c} / (2\sqrt{d})$.
Using
$\log(\sinh(\alpha)/\alpha) = \alpha^2/6 + O(\alpha^4)$,
and summing over channels:
\begin{equation}
  b_i
  \approx \sum_{c=1}^{d} \frac{\alpha_c^2}{6}
  = \frac{1}{24\,d}
    \sum_{c=1}^{d} q_c^2 \, \Delta_{i,c}^2.
  \label{eq:app_taylor}
\end{equation}
 
Under group-wise per-token quantization, where each token's $d$ channels are divided into $G = d/g$ groups sharing a common step size $\Delta_{i,j}$, this simplifies to $b_i \approx \frac{1}{24d} \sum_{j=1}^{G} \Delta_{i,j}^2 \, \|q_j\|^2$ as in \cref{eq:grouped_correction}.

\begin{figure}[h]
  \centering
  \includegraphics[width=0.65\textwidth]{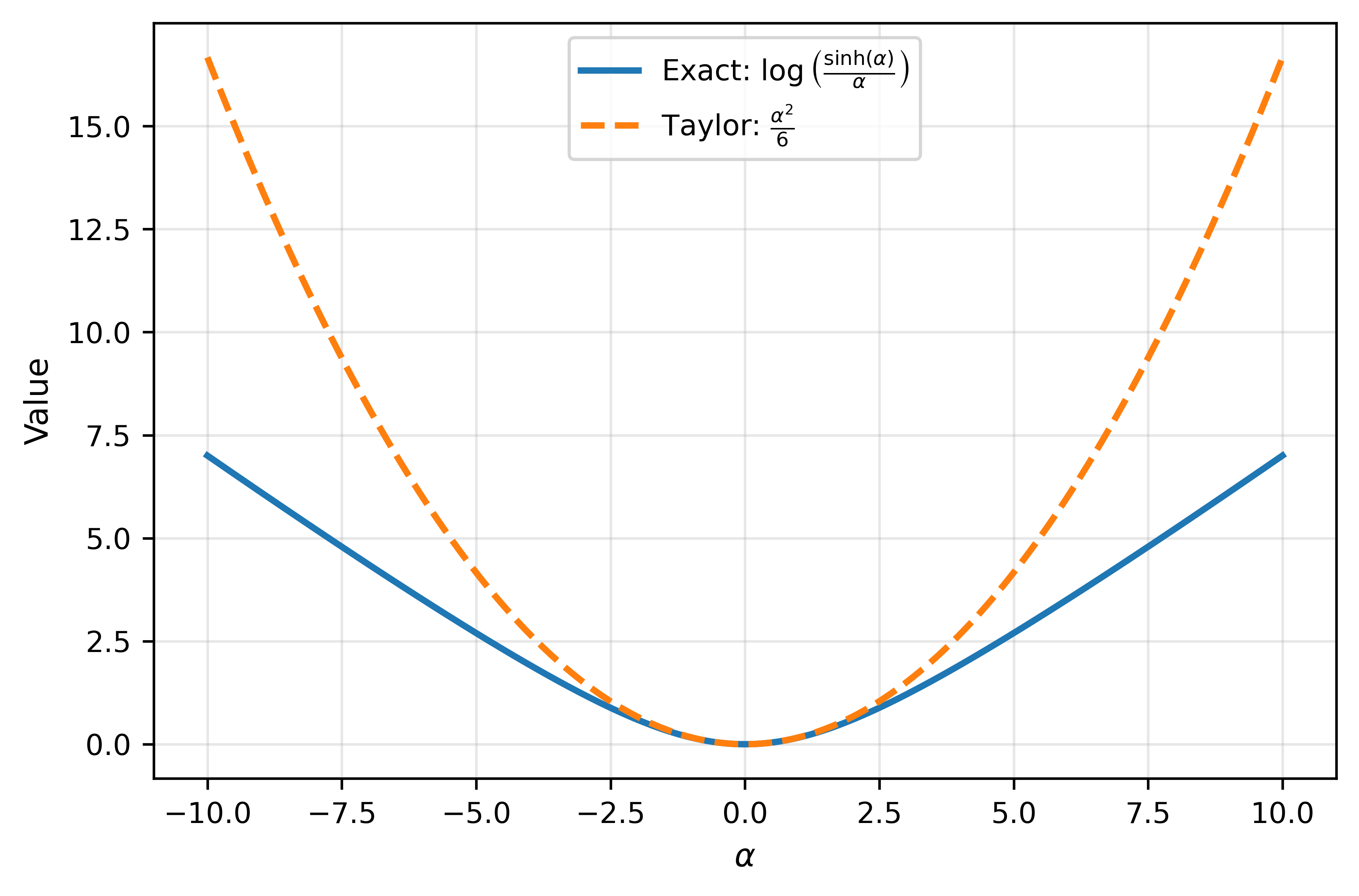}
  \caption{Exact correction $\log(\sinh(\alpha)/\alpha)$ versus its second-order Taylor approximation $\alpha^2/6$. The approximation is tight for small $|\alpha|$ but overestimates the correction for large $|\alpha|$, explaining the mild overcorrection observed at aggressive bitwidths.}
  \label{fig:exact_vs_taylor}
\end{figure}

\Cref{fig:exact_vs_taylor} compares the exact correction $\log(\sinh(\alpha)/\alpha)$ with its Taylor approximation $\alpha^2/6$ as a function of $\alpha_c = q_c \, \Delta_{i,c} / (2\sqrt{d})$. The two agree closely for small $|\alpha_c|$, but the Taylor term grows as $\alpha_c^2$ whereas the exact correction grows only as $|\alpha_c|$ for large arguments, so the approximation systematically overestimates the correction when the score-space noise is large.

At aggressive bitwidths (e.g., INT2), the approximation
may overcorrect, but we find empirically that this generally does
not harm end-to-end video quality
(see \cref{sec:experiments}).

\section{Detailed Cost Breakdown}
\label{app:detailed_cost_breakdown}

We detail the per-query, per-key, per-score-entry and total costs for the Taylor correction under group-wise per-token quantization with $G = d/g$ groups.

Under group-wise quantization with $G = d/g$ groups:

\begin{itemize}
  \item \textbf{Per-query:} Compute $\|q_j\|^2 = \sum_{c \in \mathcal{G}_j} q_c^2$ for each group $j = 1, \dots, G$, costing $O(d)$.
  \item \textbf{Per-key:} Compute $\Delta_{i,j}^2 / (24d)$ for each group, costing $O(G)$ per key.
  \item \textbf{Per score entry:} Compute an inner product between the per-query vector $(\|q_j\|^2)_{j=1}^G$ and the per-key vector $(\Delta_{i,j}^2/(24d))_{j=1}^G$, costing $O(G)$.
  \item \textbf{Total:}
  \begin{equation}
  O(Q \cdot d + K \cdot G + Q \cdot K \cdot G).
  \end{equation}
  Since $G = d/g$ and $K \gg d$, the dominant term is $O(Q \cdot K \cdot d/g)$.
  Compared to the attention cost $O(Q \cdot K \cdot d)$, this is lower by a factor of $g$.
\end{itemize}

\textbf{On storage}, we note that a cached key of dimension~$d$ quantized to $B$~bits per element
with group size~$g$ requires $d \cdot B$ bits for the quantized
values, plus metadata per group: one scale stored in
FP8 E4M3 (8~bits) and one zero-point stored in BF16
(16~bits), for a total of $24$ bits per group.
With $G = d/g$ groups per token, the effective bitwidth is
\begin{equation}
  B_{\mathrm{eff}}
  = \frac{d \cdot B + 24 \cdot G}{d}
  = B + \frac{24}{g}.
  \label{eq:effective_bits}
\end{equation}
Our correction adds no storage beyond this
($\Delta_{i,j}$ is the scale itself).
For our default configuration ($d = 128$, $g = 32$), this yields
$B_{\mathrm{eff}} = 2.75$ at INT2.

\section{Implementation Note}
\label{app:flexattention_impl}

In our implementation, the correction subtracts a per-attention-score
value $b_i$ from cached scores before softmax.
Materializing this correction for every score entry would require
a dense tensor with the same shape as the full score matrix,
which is unnecessary for long contexts.
Instead, we apply the bias on the fly through a
\texttt{score\_mod} function in PyTorch's
FlexAttention~\citep{dong2024flexattention}, which lets the fused
attention kernel incorporate the correction without materializing
the full correction tensor.

All MAGI-1 experiments were conducted on NVIDIA L4 GPUs, SkyReels-V2 experiments on NVIDIA A100 GPUs, and HY-WorldPlay experiments on NVIDIA A100 80GB GPUs.

\section{Pseudocode for Taylor-Corrected Attention}
\label{app:pseudocode}

\cref{alg:taylor_corrected_attention} provides the full pseudocode for attention with the Taylor correction applied to quantized cached keys, as derived in \cref{sec:computing_correction}.

\begin{algorithm}[h]
  \caption{Attention with Taylor correction for quantized cached keys (group-wise)}
  \label{alg:taylor_corrected_attention}
  \KwIn{Query matrix $Q \in \mathbb{R}^{M \times d}$; cached quantized keys $K_{\mathcal{S}}^{q}$ with per-group step sizes $\{\Delta_{i,j}\}$; cached values $V_{\mathcal{S}}$; current-chunk keys $K_{\mathcal{R}}$; current-chunk values $V_{\mathcal{R}}$; group size $g$, number of groups $G = d/g$}
  \KwOut{Attention output $O \in \mathbb{R}^{M \times d_v}$}
  $\hat{K}_{\mathcal{S}} \gets \mathrm{dequant}(K_{\mathcal{S}}^{q})$\;
  $S_{\mathcal{S}} \gets Q \hat{K}_{\mathcal{S}}^\top / \sqrt{d}$\;
  $S_{\mathcal{R}} \gets Q K_{\mathcal{R}}^\top / \sqrt{d}$\;
  \For{$m = 1$ \KwTo $M$}{
    \For{$j = 1$ \KwTo $G$}{
      $\nu_{m,j} \gets \sum_{c \in \mathcal{G}_j} Q_{m,c}^2$\;
    }
    \ForAll{$i \in \mathcal{S}$}{
      $b_{m,i} \gets \dfrac{1}{24\,d} \sum_{j=1}^{G} \Delta_{i,j}^2 \, \nu_{m,j}$\;
      $S_{\mathcal{S}}[m,i] \gets S_{\mathcal{S}}[m,i] - b_{m,i}$\;
    }
  }
  $S \gets \mathrm{concat}(S_{\mathcal{S}}, S_{\mathcal{R}})$\;
  $P \gets \mathrm{softmax}(S)$\;
  $V \gets \mathrm{concat}(V_{\mathcal{S}}, V_{\mathcal{R}})$\;
  $O \gets P V$\;
  \Return{$O$}
\end{algorithm}

\section{Per-Channel Quantization Correction}
\label{app:per_channel}

When quantization is performed per-channel (or group-wise per-channel),
the step size $\Delta_c$ depends on channel~$c$ but is
shared across all tokens.
The noise model becomes
$\epsilon_{i,c} \sim \mathcal{U}(-\Delta_c/2,\; +\Delta_c/2)$,
independent across channels and identically distributed across
tokens for each fixed channel.

Since $\{\Delta_c\}$ do not depend on~$i$, the distribution of
$\delta_i = \sum_c q_c \, \epsilon_{i,c} / \sqrt{d}$
is the same for all cached keys $i \in \mathcal{S}$.
The correction reduces to a single scalar shared by all tokens:
\begin{equation}
  b
  = \sum_{c=1}^{d} \log\!\left(
    \frac{
      \sinh\!\left(\dfrac{q_c\,\Delta_c}{2\sqrt{d}}\right)
    }{
      \dfrac{q_c\,\Delta_c}{2\sqrt{d}}
    }
  \right),
  \label{eq:per_channel_exact}
\end{equation}
with the Taylor approximation
\begin{equation}
  b \approx \frac{1}{24\,d}\sum_{c=1}^{d} q_c^2\,\Delta_c^2.
  \label{eq:per_channel_taylor}
\end{equation}

\paragraph{Per-channel correction.}
Since $b$ is the same for all $i \in \mathcal{S}$, the corrected
scores within the cached chunk are
$\tilde{s}_i = \hat{s}_i - b$ for all $i \in \mathcal{S}$.
Subtracting $b$ from all cached scores reduces
$Z_{\mathcal{S}}$ relative to $Z_{\mathcal{R}}$, restoring
the inter-chunk attention balance.

Under per-token quantization, the correction $b_i$ varies across
tokens, allowing it to differentially adjust each token's
contribution.
In our experiments, per-token quantization with the
token-dependent correction consistently outperforms per-channel
quantization with a shared correction.

\section{Extension to QuaRot}
\label{app:quarot}

The derivation in \cref{sec:method} assumes the unrotated space. We now
extend the correction to QuaRot (see \cref{sec:bg_quarot}).

With the Hadamard matrix $H$ applied to both keys and queries,
the quantized score becomes
\begin{equation}
  \hat{s}_i
  = \frac{(Hq)^\top(Hk_i + \epsilon_i)}{\sqrt{d}}
  = s_i + \delta_i^{(H)},
  \label{eq:quarot_score}
\end{equation}
where $\delta_i^{(H)} = (Hq)^\top \epsilon_i / \sqrt{d}$.
Our correction applies identically with $q$ replaced by~$Hq$:
$b_i^{(H)} = \log \mathbb{E}[e^{\delta_i^{(H)}}]$.

\paragraph{Taylor approximation under rotation.}
The Taylor approximation replaces $\|q_j\|^2$ with
$\|(Hq)_j\|^2$ (the per-group squared norms of the rotated
query):
\begin{equation}
  b_i^{(H)}
  \approx \frac{1}{24\,d}
    \sum_{j=1}^{G} \Delta_{i,j}^2 \, \|(Hq)_j\|^2.
  \label{eq:quarot_taylor}
\end{equation}
Note that while $\|Hq\|^2 = \|q\|^2$ by orthogonality, the
per-group norms $\|(Hq)_j\|^2$ generally differ from
$\|q_j\|^2$ because Hadamard rotation mixes channels across
groups.

\section{Fidelity Metric Standard Errors}
\label{app:fidelity_se}

\Cref{tab:main_results} reports fidelity metrics (PSNR, SSIM, LPIPS)
averaged across prompts. \Cref{tab:fidelity_se} reports the same
values with standard errors computed across prompts (the independent
sampling unit), using the evaluation data described in \cref{sec:setup_exp}.

\begin{table}[t]
  \centering
  \caption{Fidelity metrics with standard errors for all configurations
    in \cref{tab:main_results}. PSNR, SSIM, and LPIPS are computed
    relative to the BF16 reference; $\pm$ denotes standard error
    across prompts. Best quantized result per model is \textbf{bolded}.}
  \label{tab:fidelity_se}
  \small
  \setlength{\tabcolsep}{3pt}
  \begin{tabular}{llccccc}
    \toprule
    \textbf{Model}
      & \shortstack{\textbf{Quant.}\\\textbf{scheme}}
      & \textbf{Prec.}
      & \shortstack{\textbf{With}\\\textbf{corr.}}
      & \textbf{PSNR}$\uparrow$
      & \textbf{SSIM}$\uparrow$
      & \textbf{LPIPS}$\downarrow$ \\
    \midrule
    \multirow{6}{*}{MAGI-1}
      & \multirow{2}{*}{RTN}
      & \multirow{2}{*}{INT2}
      & $\times$     & 23.17\se{0.24} & 0.799\se{0.006} & 0.195\se{0.006} \\
      & & & $\checkmark$ & 24.08\se{0.30} & 0.828\se{0.006} & 0.131\se{0.004} \\
    \cmidrule(lr){2-7}
      & \multirow{2}{*}{QuaRot+RTN}
      & \multirow{2}{*}{INT2}
      & $\times$     & 17.10\se{0.13} & 0.630\se{0.009} & 0.453\se{0.007} \\
      & & & $\checkmark$ & 22.97\se{0.32} & 0.801\se{0.007} & 0.165\se{0.006} \\
    \cmidrule(lr){2-7}
      & \multirow{2}{*}{QVG}
      & \multirow{2}{*}{INT2}
      & $\times$     & 23.01\se{0.27} & 0.826\se{0.006} & 0.132\se{0.004} \\
      & & & $\checkmark$ & \textbf{25.29}\se{0.32} & \textbf{0.856}\se{0.005} & \textbf{0.107}\se{0.004} \\
    \midrule
    \multirow{4}{*}{SkyReels-V2}
      & \multirow{2}{*}{RTN}
      & \multirow{2}{*}{INT2}
      & $\times$     & 18.38\se{0.16} & 0.670\se{0.007} & 0.383\se{0.007} \\
      & & & $\checkmark$ & 18.39\se{0.21} & 0.743\se{0.008} & 0.269\se{0.008} \\
    \cmidrule(lr){2-7}
      & \multirow{2}{*}{QuaRot+RTN}
      & \multirow{2}{*}{INT2}
      & $\times$     & 19.20\se{0.21} & 0.708\se{0.009} & 0.319\se{0.008} \\
      & & & $\checkmark$ & \textbf{20.42}\se{0.26} & \textbf{0.784}\se{0.009} & \textbf{0.202}\se{0.009} \\
    \midrule
    \multirow{4}{*}{HY-WorldPlay}
      & \multirow{2}{*}{RTN}
      & \multirow{2}{*}{INT2}
      & $\times$     & 16.96\se{0.66} & 0.573\se{0.034} & 0.390\se{0.030} \\
      & & & $\checkmark$ & 17.15\se{0.60} & 0.577\se{0.030} & 0.311\se{0.022} \\
    \cmidrule(lr){2-7}
      & \multirow{2}{*}{QuaRot+RTN}
      & \multirow{2}{*}{INT2}
      & $\times$     & 17.16\se{0.62} & 0.575\se{0.034} & 0.376\se{0.028} \\
      & & & $\checkmark$ & \textbf{18.27}\se{0.59} & \textbf{0.616}\se{0.030} & \textbf{0.273}\se{0.024} \\
    \bottomrule
  \end{tabular}
\end{table}

\section{Per-Dimension VBench Results}
\label{app:vbench_dimensions}

\Cref{tab:main_results} reports the aggregate VBench Score in the
VBench-Long setting from VBench++~\citep{huang2024vbenchpp} on MAGI-1
and SkyReels-V2.
For completeness,
\cref{tab:vbench_quality,tab:vbench_semantic} break this score down
across all 16 VBench dimensions, grouped by VBench's \emph{Quality}
(visual fidelity) and \emph{Semantic} (prompt fidelity) categories,
and \cref{tab:vbench_aggregate} reports the corresponding sub-scores
together with the Total VBench Score that already appears in
\cref{tab:main_results}.
All scores are reported with standard errors across prompts
($\pm$ SE); within-prompt clips are averaged before computing the SE.

% -------------------------------------------------------------------------
% Quality table
% -------------------------------------------------------------------------
\begin{table*}[t]
  \centering
  \caption{Per-dimension VBench \emph{Quality} results on MAGI-1 and
    SkyReels-V2 (subject consistency, background consistency, temporal
    flickering, motion smoothness, dynamic degree, aesthetic quality,
    imaging quality). Values are on the standard VBench 0--100 scale.
    $\pm$ denotes standard error across prompts.
    Best quantized result per model is \textbf{bolded}.}
  \label{tab:vbench_quality}
  \footnotesize
  \setlength{\tabcolsep}{2.5pt}
  \resizebox{\textwidth}{!}{%
  \begin{tabular}{llcccccccccc}
    \toprule
    \textbf{Model}
      & \shortstack{\textbf{Quant.}\\\textbf{scheme}}
      & \textbf{Prec.}
      & \shortstack{\textbf{With}\\\textbf{corr.}}
      & \textbf{Subj. Con.}$\uparrow$
      & \textbf{BG Con.}$\uparrow$
      & \textbf{Temp. Flick.}$\uparrow$
      & \textbf{Mot. Smo.}$\uparrow$
      & \textbf{Dyn. Deg.}$\uparrow$
      & \textbf{Aes. Q.}$\uparrow$
      & \textbf{Img. Q.}$\uparrow$ \\
    \midrule
    \multirow{7}{*}{MAGI-1}
      & ---
      & BF16
      &
      & 98.24\se{0.27} & 98.33\se{0.15} & 99.64\se{0.07} & 99.53\se{0.04} & 18.46\se{5.43} & 58.86\se{2.09} & 58.63\se{2.89} \\
    \cmidrule(lr){2-11}
      & \multirow{2}{*}{RTN}
      & \multirow{2}{*}{INT2}
      & $\times$
      & 97.71\se{0.31} & 98.01\se{0.07} & 99.57\se{0.07} & 99.41\se{0.05} & 15.38\se{4.34} & 55.50\se{2.12} & 54.50\se{3.09} \\
      & & & $\checkmark$
      & 98.20\se{0.23} & \textbf{98.22}\se{0.08} & \textbf{99.64}\se{0.07} & \textbf{99.54}\se{0.03} & 15.38\se{4.87} & 58.23\se{2.04} & 57.84\se{2.96} \\
    \cmidrule(lr){2-11}
      & \multirow{2}{*}{QuaRot+RTN}
      & \multirow{2}{*}{INT2}
      & $\times$
      & 94.26\se{0.24} & 96.98\se{0.09} & 99.10\se{0.14} & 98.41\se{0.23} & \textbf{48.46}\se{5.57} & 40.71\se{0.95} & 39.72\se{1.78} \\
      & & & $\checkmark$
      & 98.01\se{0.31} & 98.19\se{0.12} & 99.61\se{0.09} & 99.53\se{0.04} & 16.92\se{5.62} & 57.98\se{1.98} & 58.16\se{2.79} \\
    \cmidrule(lr){2-11}
      & \multirow{2}{*}{QVG}
      & \multirow{2}{*}{INT2}
      & $\times$
      & 97.93\se{0.29} & 98.10\se{0.11} & 99.51\se{0.09} & 99.47\se{0.04} & 17.69\se{5.12} & 57.87\se{2.08} & 57.78\se{2.85} \\
      & & & $\checkmark$
      & \textbf{98.24}\se{0.24} & \textbf{98.22}\se{0.11} & 99.57\se{0.09} & 99.52\se{0.04} & 17.69\se{5.47} & \textbf{58.60}\se{2.05} & \textbf{58.63}\se{2.85} \\
    \midrule
    \multirow{5}{*}{SkyReels-V2}
      & ---
      & BF16
      &
      & 97.66\se{0.30} & 97.46\se{0.13} & 99.57\se{0.09} & 99.15\se{0.09} & 79.81\se{6.05} & 53.87\se{2.15} & 59.04\se{2.18} \\
    \cmidrule(lr){2-11}
      & \multirow{2}{*}{RTN}
      & \multirow{2}{*}{INT2}
      & $\times$
      & 92.69\se{0.37} & 95.75\se{0.15} & 99.15\se{0.05} & 98.04\se{0.24} & 33.65\se{6.79} & 41.23\se{1.65} & 45.20\se{2.52} \\
      & & & $\checkmark$
      & 96.98\se{0.40} & 96.60\se{0.18} & \textbf{99.53}\se{0.06} & 98.94\se{0.09} & \textbf{85.58}\se{5.22} & \textbf{56.06}\se{1.95} & \textbf{63.38}\se{2.14} \\
    \cmidrule(lr){2-11}
      & \multirow{2}{*}{QuaRot+RTN}
      & \multirow{2}{*}{INT2}
      & $\times$
      & 94.13\se{0.38} & 95.91\se{0.18} & 99.34\se{0.07} & 98.47\se{0.18} & 51.92\se{7.96} & 41.33\se{1.73} & 48.43\se{2.57} \\
      & & & $\checkmark$
      & \textbf{97.56}\se{0.29} & \textbf{97.05}\se{0.17} & 99.50\se{0.10} & \textbf{99.09}\se{0.09} & 81.73\se{5.81} & 52.49\se{2.06} & 58.22\se{2.35} \\
    \bottomrule
  \end{tabular}%
  }
\end{table*}

% -------------------------------------------------------------------------
% Semantic table
% -------------------------------------------------------------------------
\begin{table*}[t]
  \centering
  \caption{Per-dimension VBench \emph{Semantic} results on MAGI-1 and
    SkyReels-V2 (object class, multiple objects, human action, color,
    spatial relationship, scene, appearance style, temporal style,
    overall consistency). Values are on the standard VBench 0--100
    scale. $\pm$ denotes standard error across prompts.
    Best quantized result per model is \textbf{bolded};
    ties are bolded jointly.}
  \label{tab:vbench_semantic}
  \footnotesize
  \setlength{\tabcolsep}{2pt}
  \resizebox{\textwidth}{!}{%
  \begin{tabular}{llcccccccccccc}
    \toprule
    \textbf{Model}
      & \shortstack{\textbf{Quant.}\\\textbf{scheme}}
      & \textbf{Prec.}
      & \shortstack{\textbf{With}\\\textbf{corr.}}
      & \textbf{Obj. Cls.}$\uparrow$
      & \textbf{Mult. Obj.}$\uparrow$
      & \textbf{Hum. Act.}$\uparrow$
      & \textbf{Color}$\uparrow$
      & \textbf{Spat. Rel.}$\uparrow$
      & \textbf{Scene}$\uparrow$
      & \textbf{App. Sty.}$\uparrow$
      & \textbf{Temp. Sty.}$\uparrow$
      & \textbf{Overall Con.}$\uparrow$ \\
    \midrule
    \multirow{7}{*}{MAGI-1}
      & ---
      & BF16
      &
      & 100.00\se{0.00} & 57.40\se{9.18} & 80.77\se{7.88} & 96.00\se{4.00} & 71.30\se{7.61} & 18.70\se{7.16} & 22.42\se{0.69} & 22.38\se{0.59} & 25.53\se{1.14} \\
    \cmidrule(lr){2-13}
      & \multirow{2}{*}{RTN}
      & \multirow{2}{*}{INT2}
      & $\times$
      & \textbf{100.00}\se{0.00} & 51.44\se{9.28} & 76.92\se{8.43} & 95.33\se{4.02} & 69.10\se{7.41} & 21.20\se{7.14} & \textbf{23.04}\se{0.66} & 20.83\se{0.62} & 25.11\se{1.20} \\
      & & & $\checkmark$
      & \textbf{100.00}\se{0.00} & 55.10\se{9.16} & 80.77\se{7.88} & 95.55\se{4.01} & 68.80\se{8.19} & \textbf{21.92}\se{7.72} & 22.53\se{0.71} & \textbf{22.48}\se{0.60} & 25.48\se{1.15} \\
    \cmidrule(lr){2-13}
      & \multirow{2}{*}{QuaRot+RTN}
      & \multirow{2}{*}{INT2}
      & $\times$
      & 43.17\se{3.59} & 18.85\se{3.65} & \textbf{88.46}\se{6.39} & 88.75\se{4.14} & 25.41\se{3.14} & 8.94\se{3.30} & 23.00\se{0.37} & 16.84\se{0.85} & 22.67\se{1.19} \\
      & & & $\checkmark$
      & \textbf{100.00}\se{0.00} & \textbf{55.48}\se{8.73} & 84.62\se{7.22} & \textbf{96.77}\se{4.00} & \textbf{70.43}\se{8.03} & 20.82\se{7.20} & 22.38\se{0.71} & 22.37\se{0.60} & 25.26\se{1.22} \\
    \cmidrule(lr){2-13}
      & \multirow{2}{*}{QVG}
      & \multirow{2}{*}{INT2}
      & $\times$
      & \textbf{100.00}\se{0.00} & 55.34\se{9.28} & 84.62\se{7.22} & 95.25\se{4.04} & 67.81\se{7.82} & 19.90\se{7.19} & 22.66\se{0.67} & 22.10\se{0.59} & 25.43\se{1.18} \\
      & & & $\checkmark$
      & \textbf{100.00}\se{0.00} & 55.05\se{9.15} & 84.62\se{7.22} & \textbf{96.77}\se{4.00} & 70.16\se{7.57} & 21.25\se{7.60} & 22.46\se{0.71} & 22.22\se{0.58} & \textbf{25.51}\se{1.14} \\
    \midrule
    \multirow{5}{*}{SkyReels-V2}
      & ---
      & BF16
      &
      & 75.30\se{8.08} & 51.74\se{8.60} & 76.92\se{8.43} & 77.97\se{7.40} & 67.41\se{8.48} & 12.44\se{6.31} & 18.86\se{0.56} & 18.93\se{0.98} & 21.01\se{1.27} \\
    \cmidrule(lr){2-13}
      & \multirow{2}{*}{RTN}
      & \multirow{2}{*}{INT2}
      & $\times$
      & 46.69\se{6.93} & 18.51\se{3.77} & 65.38\se{9.51} & 71.98\se{5.48} & 34.87\se{6.01} & 6.67\se{3.24} & \textbf{20.66}\se{0.42} & 20.10\se{0.61} & 22.34\se{0.95} \\
      & & & $\checkmark$
      & \textbf{75.96}\se{7.30} & 50.78\se{8.45} & \textbf{80.77}\se{7.88} & \textbf{77.71}\se{6.75} & 68.97\se{7.86} & \textbf{14.06}\se{6.61} & 18.82\se{0.55} & \textbf{20.34}\se{0.78} & \textbf{22.73}\se{1.08} \\
    \cmidrule(lr){2-13}
      & \multirow{2}{*}{QuaRot+RTN}
      & \multirow{2}{*}{INT2}
      & $\times$
      & 59.68\se{8.18} & 28.31\se{5.15} & 61.54\se{9.73} & 75.06\se{6.10} & 45.87\se{7.62} & 11.72\se{5.09} & 19.86\se{0.54} & 18.53\se{0.84} & 20.54\se{1.31} \\
      & & & $\checkmark$
      & 73.26\se{8.11} & \textbf{52.88}\se{8.40} & 76.92\se{8.43} & 74.62\se{7.45} & \textbf{69.00}\se{7.63} & 13.46\se{6.54} & 18.82\se{0.56} & 19.35\se{0.77} & 20.82\se{1.27} \\
    \bottomrule
  \end{tabular}%
  }
\end{table*}

% -------------------------------------------------------------------------
% Aggregate table
% -------------------------------------------------------------------------
\begin{table*}[t]
  \centering
  \caption{Aggregate VBench scores for MAGI-1 and SkyReels-V2:
    VBench's Quality and Semantic sub-scores and the total VBench
    Score (which already appears in \cref{tab:main_results}).
    Values are on the standard VBench 0--100 scale.
    Best quantized result per model is \textbf{bolded}.
    $\pm$ denotes standard error across prompts, propagated to
    aggregate scores via linear error propagation through VBench's
    normalization and weighting.}
  \label{tab:vbench_aggregate}
  \small
  \setlength{\tabcolsep}{3pt}
  \begin{tabular}{llccccc}
    \toprule
    \textbf{Model}
      & \shortstack{\textbf{Quant.}\\\textbf{scheme}}
      & \textbf{Prec.}
      & \shortstack{\textbf{With}\\\textbf{corr.}}
      & \textbf{Quality}$\uparrow$
      & \textbf{Semantic}$\uparrow$
      & \textbf{Total}$\uparrow$ \\
    \midrule
    \multirow{7}{*}{MAGI-1}
      & ---
      & BF16
      &
      & 80.10\se{0.69} & 70.93\se{1.97} & 78.27\se{0.68} \\
    \cmidrule(lr){2-7}
      & \multirow{2}{*}{RTN}
      & \multirow{2}{*}{INT2}
      & $\times$     & 78.46\se{0.67} & 69.49\se{2.00} & 76.67\se{0.67} \\
      & & & $\checkmark$ & 79.62\se{0.67} & 70.83\se{2.04} & 77.86\se{0.67} \\
    \cmidrule(lr){2-7}
      & \multirow{2}{*}{QuaRot+RTN}
      & \multirow{2}{*}{INT2}
      & $\times$     & 74.90\se{0.55} & 51.62\se{1.26} & 70.24\se{0.50} \\
      & & & $\checkmark$ & 79.69\se{0.68} & 71.31\se{1.94} & 78.02\se{0.67} \\
    \cmidrule(lr){2-7}
      & \multirow{2}{*}{QVG}
      & \multirow{2}{*}{INT2}
      & $\times$     & 79.57\se{0.67} & 70.79\se{1.96} & 77.81\se{0.67} \\
      & & & $\checkmark$ & \textbf{79.95}\se{0.69} & \textbf{71.35}\se{1.97} & \textbf{78.23}\se{0.68} \\
    \midrule
    \multirow{5}{*}{SkyReels-V2}
      & ---
      & BF16
      &
      & 83.60\se{0.67} & 60.02\se{2.27} & 78.89\se{0.70} \\
    \cmidrule(lr){2-7}
      & \multirow{2}{*}{RTN}
      & \multirow{2}{*}{INT2}
      & $\times$     & 73.97\se{0.71} & 48.27\se{1.74} & 68.83\se{0.67} \\
      & & & $\checkmark$ & \textbf{84.62}\se{0.61} & \textbf{61.71}\se{2.15} & \textbf{80.04}\se{0.65} \\
    \cmidrule(lr){2-7}
      & \multirow{2}{*}{QuaRot+RTN}
      & \multirow{2}{*}{INT2}
      & $\times$     & 76.48\se{0.79} & 51.28\se{2.06} & 71.44\se{0.75} \\
      & & & $\checkmark$ & 83.25\se{0.66} & 59.91\se{2.24} & 78.58\se{0.69} \\
    \bottomrule
  \end{tabular}
\end{table*}

\section{Qualitative Comparison on SkyReels-V2}
\label{app:qualitative_skyreels}

\begin{figure}[t]
  \centering
  \begin{subfigure}[t]{\textwidth}
    \centering
    \includegraphics[width=\textwidth]{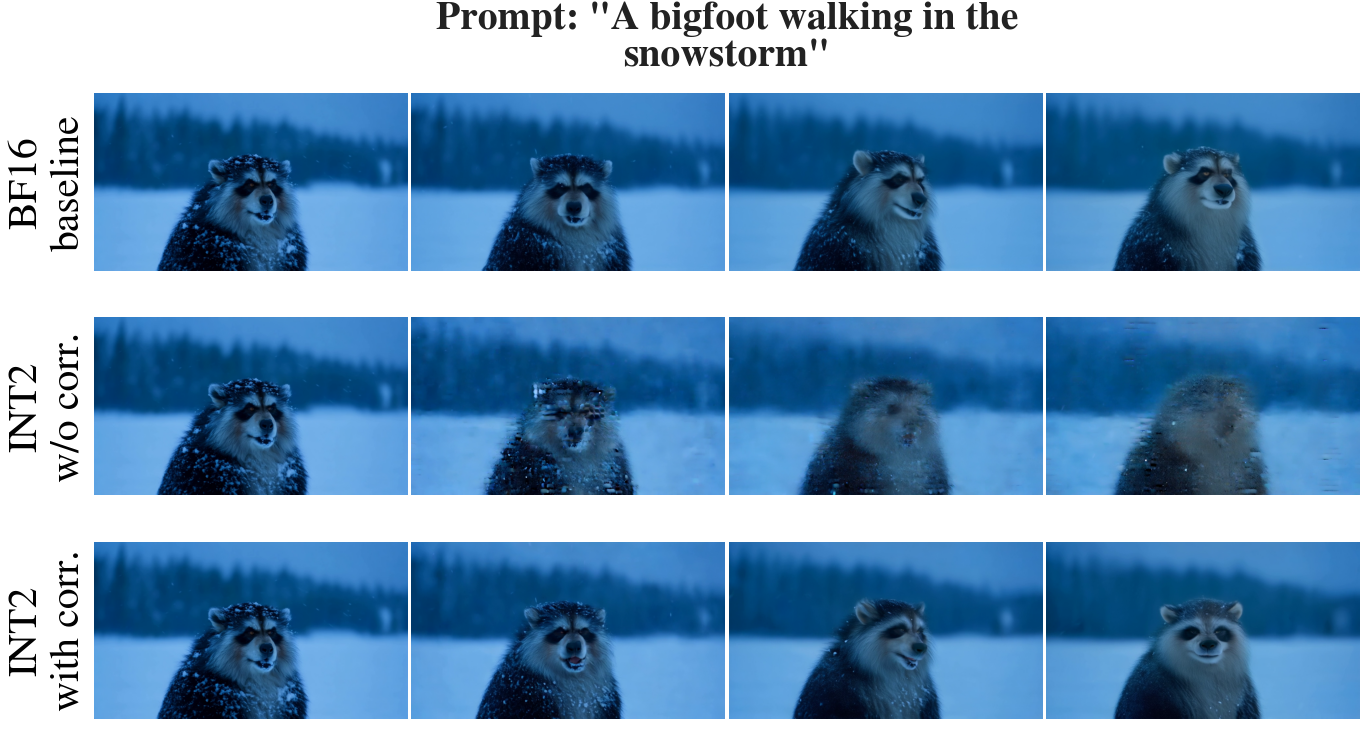}
  \end{subfigure}

  \vspace{0.5em}

  \begin{subfigure}[t]{\textwidth}
    \centering
    \includegraphics[width=\textwidth]{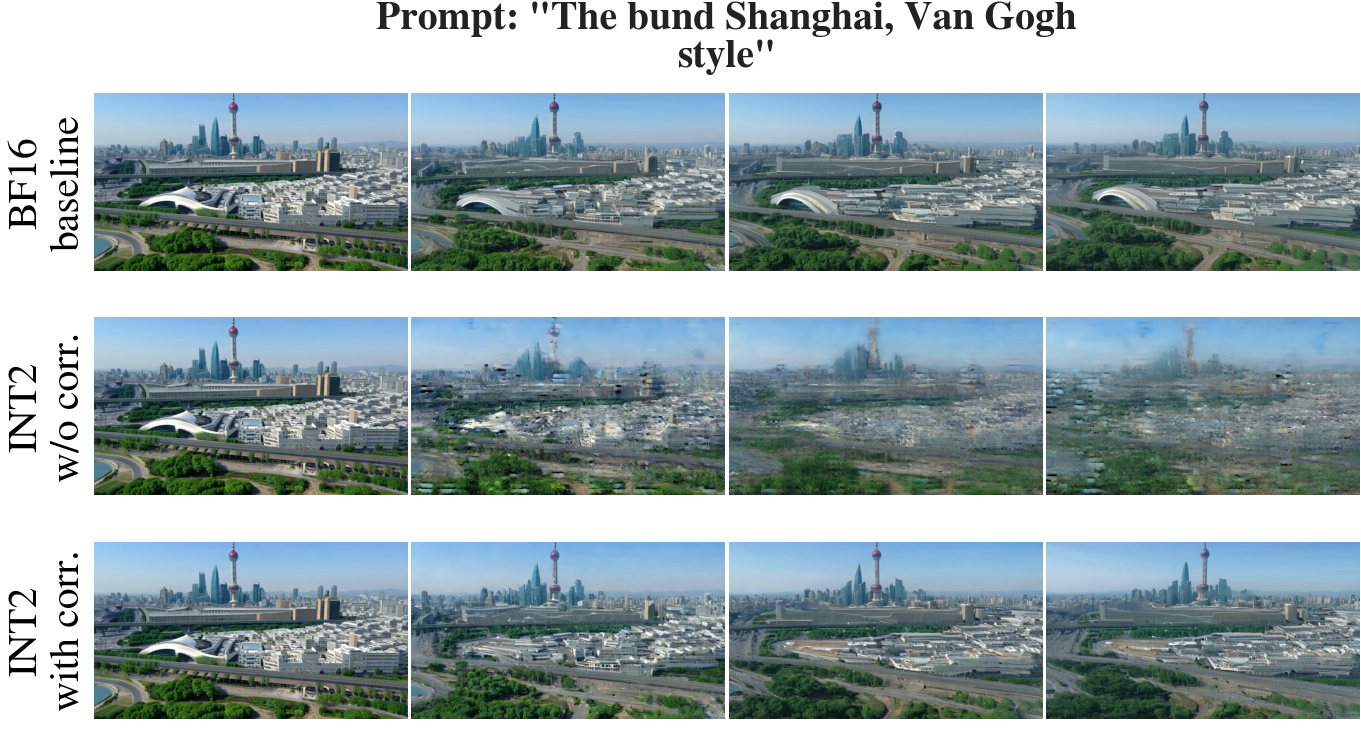}
  \end{subfigure}

  \caption{Qualitative comparison on SkyReels-V2.
    Columns show successive frames from the same video.
    Rows show BF16; INT2 asymmetric QuaRot+RTN quantization of cached
    keys and values; and the same setting with our correction.
    As on MAGI-1 (\cref{fig:qualitative_comparison}), INT2 introduces
    visible distortions, while our correction recovers much of the
    BF16-like visual quality and temporal consistency.}
  \label{fig:qualitative_comparison_skyreels}
\end{figure}

\Cref{fig:qualitative_comparison} in the main text shows the qualitative
effect of INT2 KV-cache quantization and our correction on MAGI-1.
\Cref{fig:qualitative_comparison_skyreels} reports the analogous
comparison on SkyReels-V2 for two representative prompts from the
VBench-Long suite.

\FloatBarrier

\section{Qualitative Comparison on HY-WorldPlay}
\label{app:qualitative_hywp}

\begin{figure}[!t]
  \centering
  \begin{subfigure}[t]{\textwidth}
    \centering
    \includegraphics[width=\textwidth]{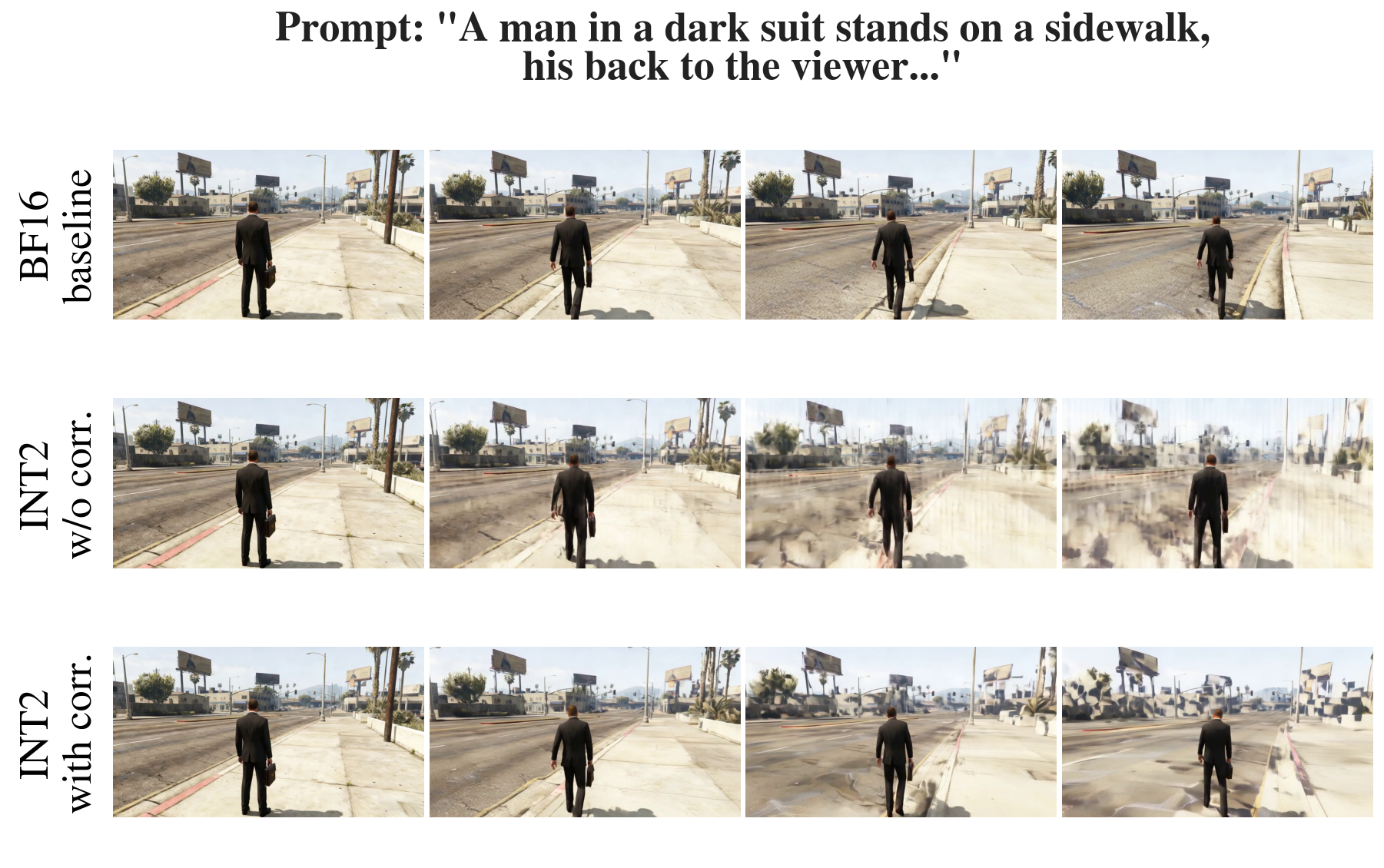}
  \end{subfigure}

  \vspace{0.5em}

  \begin{subfigure}[t]{\textwidth}
    \centering
    \includegraphics[width=\textwidth]{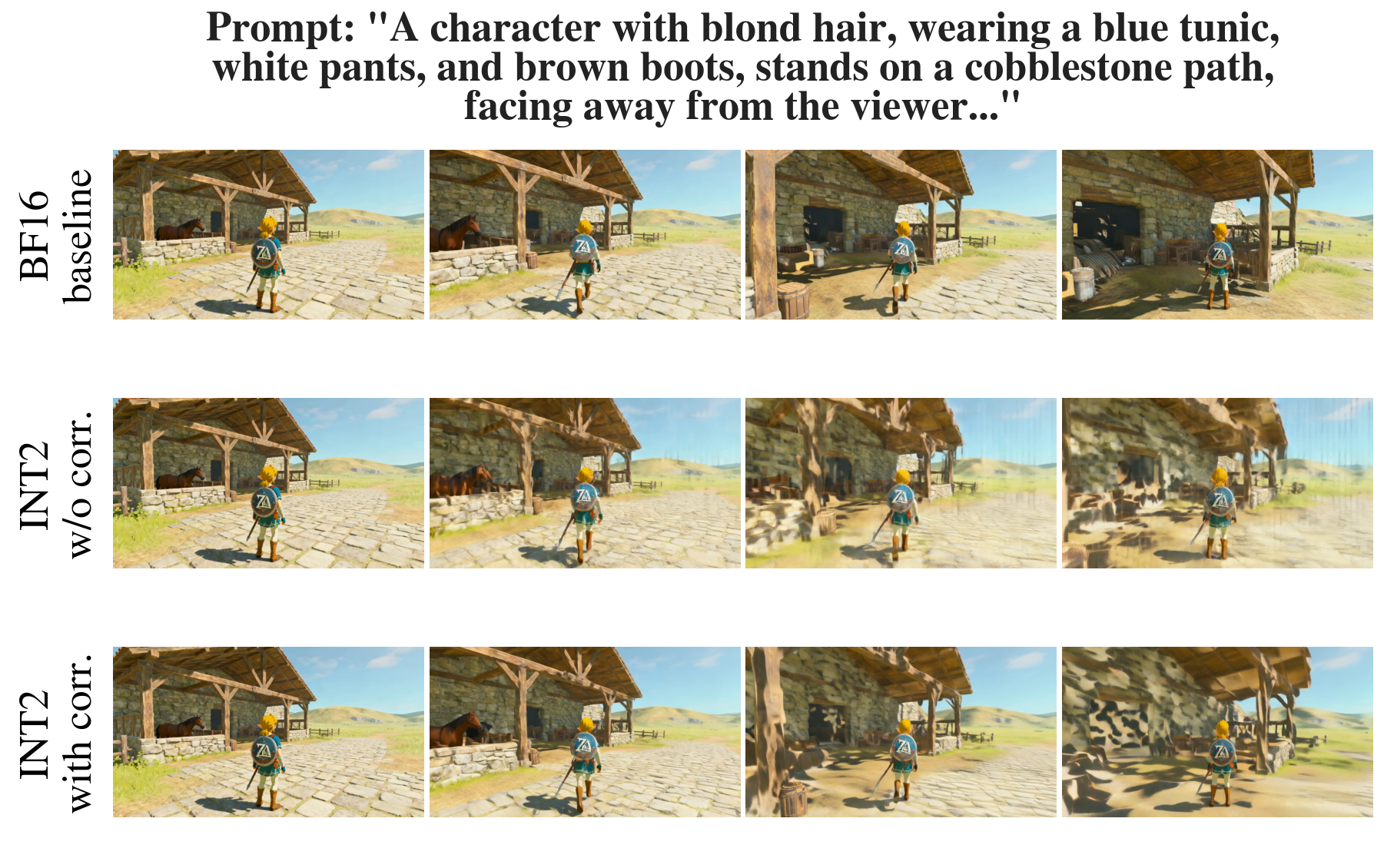}
  \end{subfigure}

  \caption{Qualitative comparison on HY-WorldPlay.
    Columns show successive frames from the same video.
    Rows show BF16; INT2 asymmetric QuaRot+RTN KV-cache
    quantization of keys and values; and the same quantized setting
    with our correction. As on MAGI-1
    (\cref{fig:qualitative_comparison}), INT2 introduces visible
    distortions, while our correction recovers much of the BF16-like
    visual quality and temporal consistency.}
  \label{fig:qualitative_comparison_hywp}
\end{figure}

\Cref{fig:qualitative_comparison} in the main text shows the qualitative
effect of INT2 KV-cache quantization and our correction on MAGI-1.
For completeness, \cref{fig:qualitative_comparison_hywp} reports the
analogous comparison on HY-WorldPlay for two representative
image--prompt pairs from the original HY-WorldPlay repository.

\FloatBarrier

\section{Attention Mass Shift}
\label{app:prob_mass_shift}

\Cref{fig:partition_sum} in the main text reports the cached
attention mass shift $\Delta P_{\mathcal{S}}$ at INT2.
For completeness, we report here the same analysis at INT4 on
MAGI-1 with the same quantization scheme and our correction.

\begin{figure}[t]
  \centering
  \includegraphics[width=0.8\textwidth]{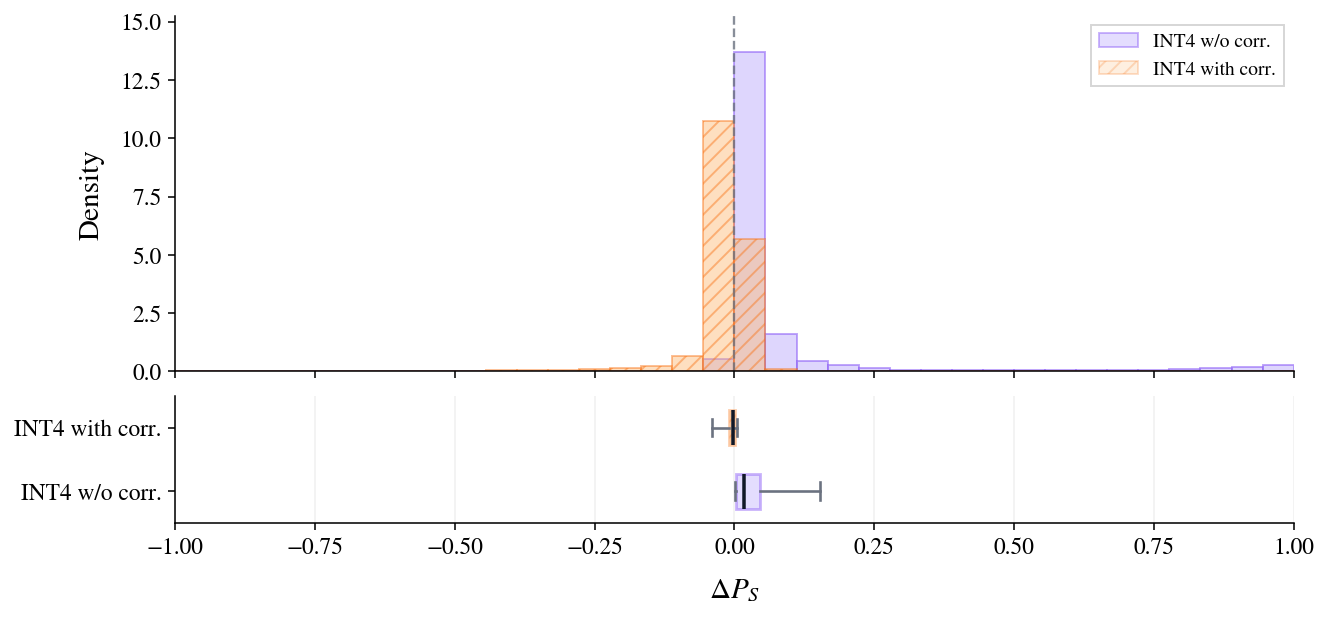}
  \caption{Cached attention mass shift $\Delta P_{\mathcal{S}}$ on
    MAGI-1 at INT4 with QuaRot+RTN KV-cache quantization.
    The same qualitative pattern as at INT2
    (cf.\ \cref{fig:partition_sum}) is visible, but the bias is
    much smaller. The correction centers the distribution near zero.}
  \label{fig:int4_partition_sum}
\end{figure}

The INT4 results in \cref{fig:int4_partition_sum} show the same
qualitative pattern as at INT2: a right-skewed quantized
distribution of $\Delta P_{\mathcal{S}}$ that the correction
centers near zero. However, the magnitude of the bias is much
smaller. Because the uncorrected bias is already small at INT4 and
generated videos are visually close to the BF16 baseline, the
correction's benefit is correspondingly mild, which is why we focus
the main paper on INT2.

\section{Attention JSD Distributions}
\label{app:jsd_distributions}

\begin{figure}[t]
  \centering
  \begin{subfigure}[t]{0.48\textwidth}
    \centering
    \includegraphics[width=\textwidth]{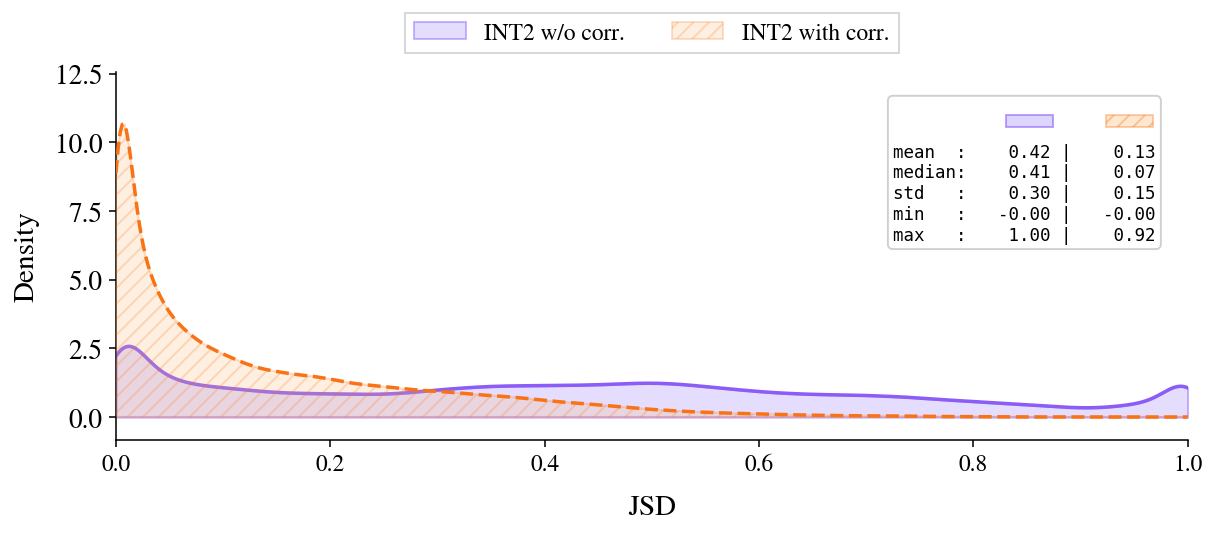}
    \caption{INT2 quantization}
  \end{subfigure}
  \hfill
  \begin{subfigure}[t]{0.48\textwidth}
    \centering
    \includegraphics[width=\textwidth]{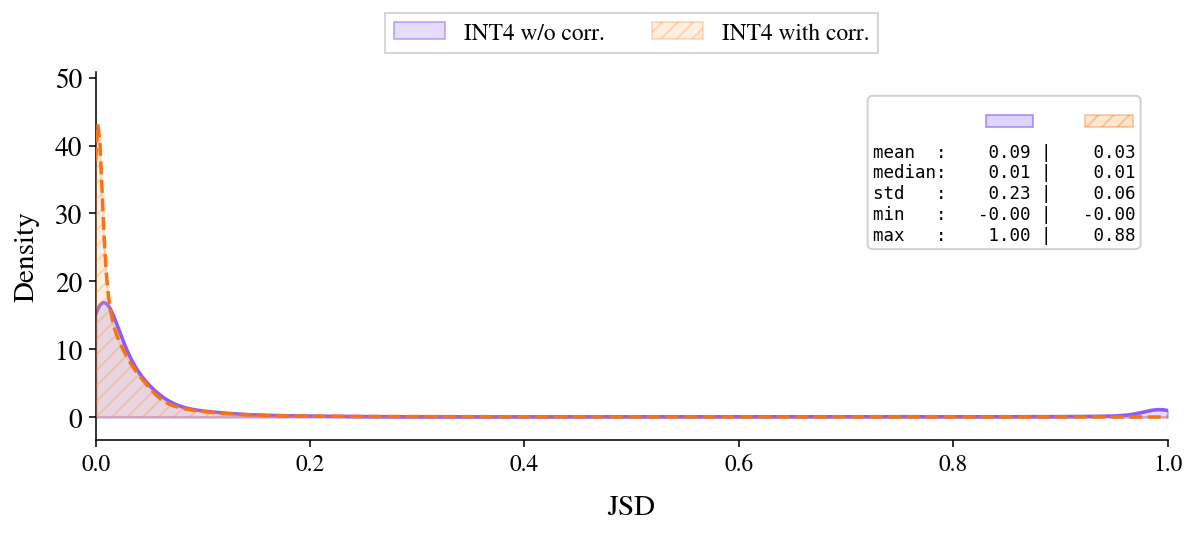}
    \caption{INT4 quantization}
  \end{subfigure}
  \caption{Distribution of Jensen-Shannon divergence between quantized
    (or corrected) and BF16 attention weights on MAGI-1 under
    QuaRot+RTN. At INT2 the correction substantially reduces the JSD;
    at INT4 the baseline JSD is already low and the improvement is modest.}
  \label{fig:jsd_distributions}
\end{figure}

\Cref{fig:jsd_distributions} plots the distribution of
Jensen-Shannon divergence (JSD) between the quantized (or corrected)
and BF16 attention weights on MAGI-1 under QuaRot+RTN quantization,
computed over all keys.
At INT2, the correction consistently shifts the JSD distribution
toward lower values, confirming that removing the partition sum bias
improves the overall attention distribution.
At INT4 the JSD is already low without correction, and the
correction provides only a modest further reduction, mirroring the
smaller probability-mass bias observed in
\cref{app:prob_mass_shift}.

\section{Attention Output MSE}
\label{app:attn_mse}

\begin{figure}[t]
  \centering
  \begin{subfigure}[t]{0.48\textwidth}
    \includegraphics[width=\textwidth]{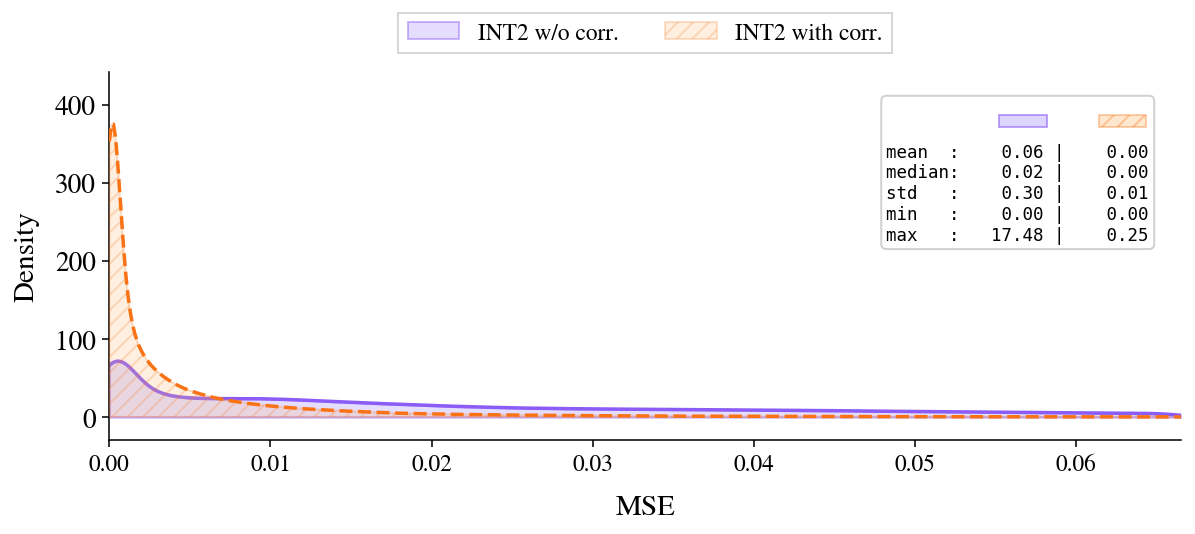}
    \caption{INT2 quantization}
  \end{subfigure}
  \hfill
  \begin{subfigure}[t]{0.48\textwidth}
    \includegraphics[width=\textwidth]{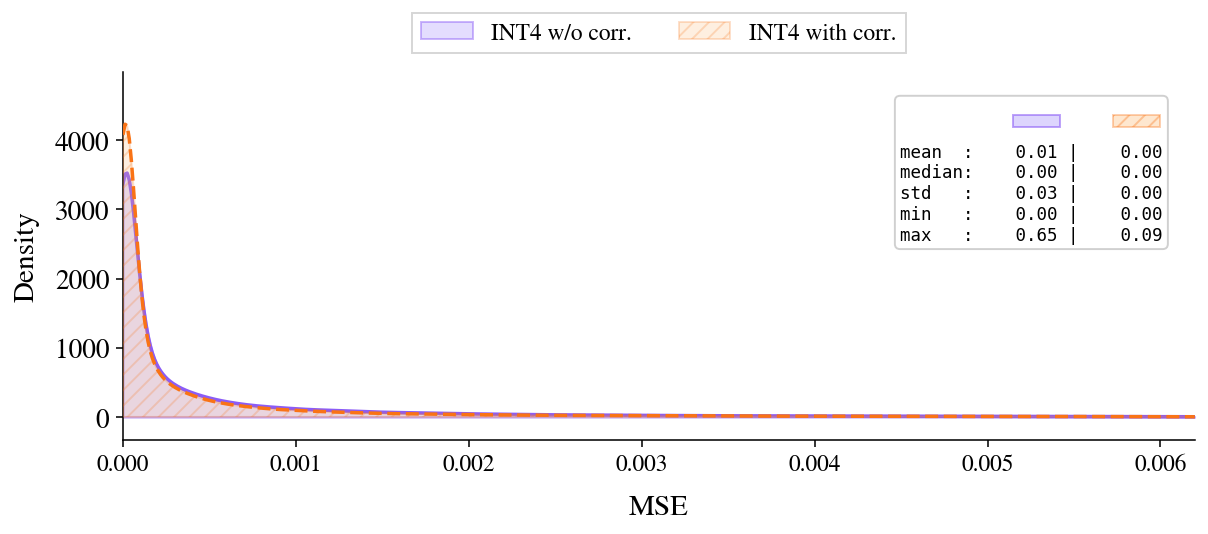}
    \caption{INT4 quantization}
  \end{subfigure}
  \caption{Attention output MSE between quantized (or corrected)
    and BF16 computations on MAGI-1 under QuaRot+RTN. The correction
    reduces MSE at INT2, confirming that score-level improvements
    propagate to the attention output. At INT4 the effect is smaller.}
  \label{fig:attn_mse}
\end{figure}

\Cref{fig:attn_mse} reports the mean squared error (MSE) of the
attention output $\mathrm{softmax}(S)\,V$ between the quantized
(or corrected) and BF16 computations on MAGI-1 under QuaRot+RTN
quantization.
At INT2, the correction consistently reduces the attention output
MSE, confirming that improvements at the score level propagate to
the attention output.
At INT4 the MSE follows the same trend as the JSD
(\cref{app:jsd_distributions}): already low without correction,
with a modest further reduction after correction.

\section{Storage--Quality Trade-Off: SSIM and LPIPS}
\label{app:storage_quality_ssim_lpips}

\begin{figure}[t]
  \centering
  \includegraphics[width=0.7\textwidth]{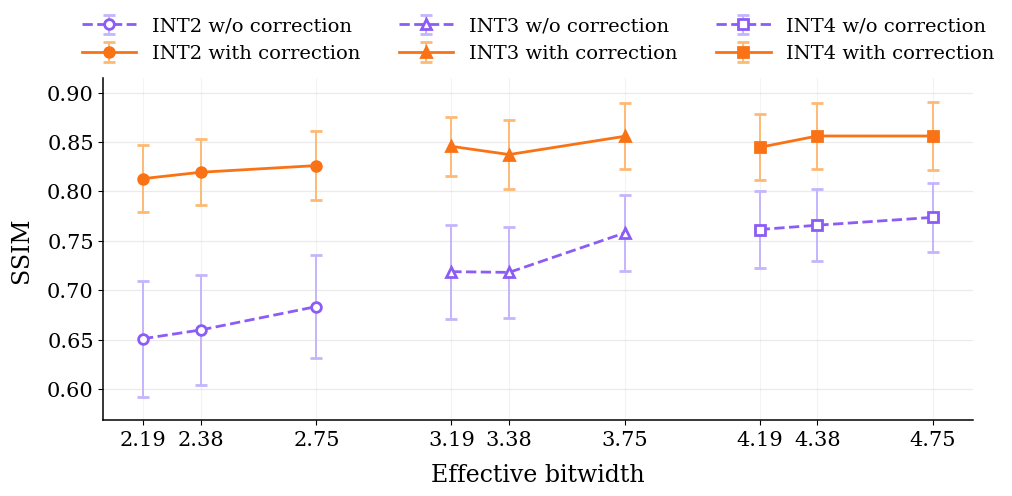}
  \caption{Trade-off between SSIM and effective bitwidth on
    MAGI-1. Same setting as
    \cref{fig:effective_bitwidth_vs_quality}.}
  \label{fig:storage_ssim}
\end{figure}

\begin{figure}[t]
  \centering
  \includegraphics[width=0.7\textwidth]{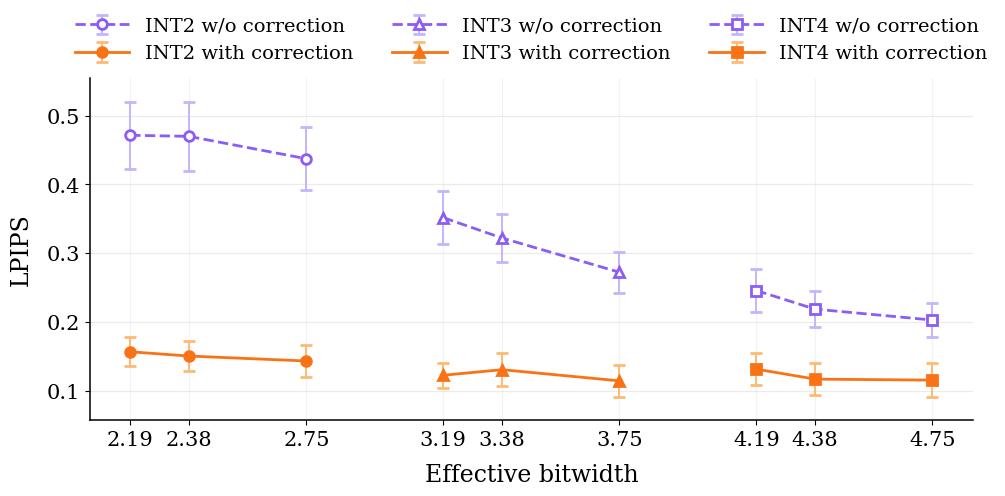}
  \caption{Trade-off between LPIPS and effective bitwidth on
    MAGI-1. Same setting as
    \cref{fig:effective_bitwidth_vs_quality}.}
  \label{fig:storage_lpips}
\end{figure}

\Cref{fig:effective_bitwidth_vs_quality} in the main text
reports the storage--quality trade-off in terms of PSNR.
For completeness, \cref{fig:storage_ssim,fig:storage_lpips}
report the same analysis for SSIM and LPIPS, confirming that
the correction uniformly improves the trade-off across all
three fidelity metrics.

\FloatBarrier

\section{LLM Partial-Prefill Experiments}
\label{app:llm_partial_prefill}

Our main experiments focus on chunk-wise autoregressive video diffusion,
where previously generated chunks are stored in a quantized KV cache and
the current chunk remains in full precision. In this appendix, we evaluate
whether our correction transfers to decoder-only language models under structurally analogous partial prefill.

Following the notation of \cref{sec:bias}, each prompt contains a quantized cached prefix $\mathcal{S}$ and a full-precision current
prefill chunk $\mathcal{R}$, with lengths $|\mathcal{S}|=A$ and $|\mathcal{R}|=B$, where $B \gg 1$.
This setup preserves the key structural feature of chunk-wise video generation: a quantized cached block $\mathcal{S}$ competes inside the same softmax with a multi-token full-precision current block $\mathcal{R}$.

These experiments provide a cross-domain validation of the bias-correction mechanism derived in Section~\ref{sec:method}, rather than a comprehensive LLM inference benchmark.

\subsection{Experimental setup}
\label{app:llm_setup}
We evaluate three decoder-only LLMs: Llama-3.1-8B~\citep{dubey2024llama3herd,meta2024llama31modelcard}, Mistral-7B-Instruct-v0.3~\citep{jiang2023mistral7b,mistralai2024mistral7binstructv03}, and Qwen2.5-32B-Instruct~\citep{qwen2024qwen25technicalreport,qwen2024qwen2532binstruct}. We use English prompts from LongBench-Pro~\citep{chen2026longbenchprorealisticcomprehensive}. We define retained prompt-length bins, e.g., $[256,512)$, $[512,1024)$, etc., then deterministically truncate prompts to retained lengths sampled uniformly from the corresponding
bin. Each evaluation job uses one fixed current-chunk size across the resulting mixed prompt lengths.

For each model and chunk size, we use the same INT2 KV-cache quantization as in the main paper. We apply our Taylor-approximated score correction to cached-key attention scores before softmax, as described in \ref{sec:method}.

Completed runs cover current-chunk sizes from $128$ to $8192$; larger
attempted configurations exceeded accelerator memory even on 80~GB GPUs.
This is due to the quadratic workspace of partial-prefill attention, whose
dense score tensor scales as $H B(A+B)$, where $H$ is the number of
attention heads, $A=|\mathcal{S}|$ is the cached-prefix length, and $B=|\mathcal{R}|$ is the current-chunk length.
To avoid artifacts from this missingness, all aggregate results are reported
as paired comparisons: each difference is computed only within cells matched
by model, current-chunk size, prompt-length bin, and evaluation examples.

Our primary metric is teacher-forced negative log-likelihood (NLL). For a set of evaluation examples $\mathcal{D}$, we aggregate at corpus level:
\[
\mathrm{NLL}
=
\frac{
\sum_{x \in \mathcal{D}} \sum_{t=1}^{T_x}
-\log p_{\theta}(y_t \mid y_{<t}, x)
}{
\sum_{x \in \mathcal{D}} T_x
}.
\]
We use NLL as the main metric because it aggregates token-level likelihoods directly and avoids the heavy-tailed behavior of averaging per-example perplexities.

\subsection{LLM partial prefill results}
\label{app:llm_aggregate_results}

\begin{figure}[t]
  \centering
  \includegraphics[width=\textwidth]{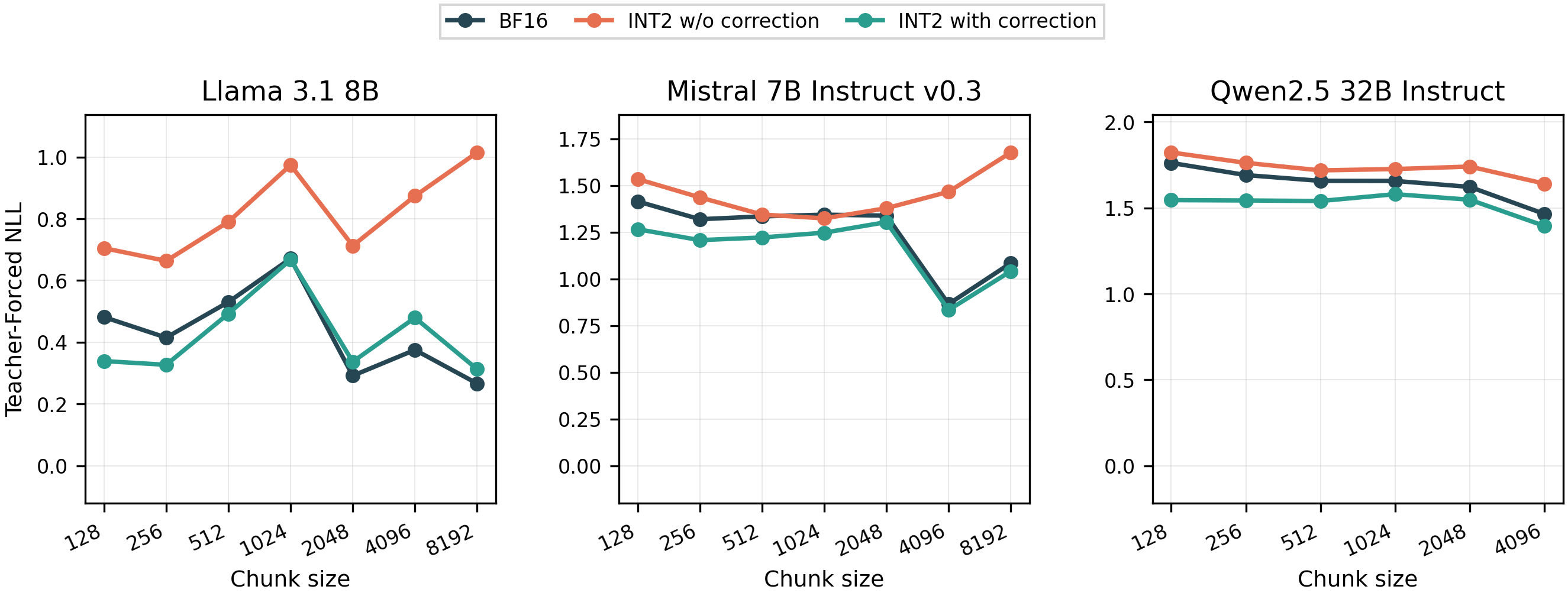}
  \caption{
  Teacher-forced NLL by partial-prefill chunk size in the LLM partial-prefill setting. Each panel corresponds to one model, and curves show BF16, plain INT2 KV-cache quantization, and INT2 with Taylor correction. Plain INT2 generally increases NLL, while the Taylor correction consistently reduces the degradation.
  }
  \label{fig:llm_nll_by_chunk}
\end{figure}

Figure~\ref{fig:llm_nll_by_chunk} summarizes our findings for the LLM ablation study. Plain INT2 KV-cache quantization consistently worsens teacher-forced NLL, while the Taylor correction improves over plain INT2 across the completed model and chunk-size settings. The corrected condition is sometimes below the BF16 NLL, although we interpret this conservatively as a partial-prefill rebalancing effect rather than as evidence that the method generally improves over full precision.

We observe substantial degradation from INT2 KV-cache quantization, especially at large chunk sizes for the smaller Mistral-7B-Instruct-v0.3 and Llama-3.2-1B models. The larger Qwen2.5 model shows smaller plain-INT2 degradation, but the correction still consistently improves NLL. This suggests that the correction is useful both in severe degradation regimes and in milder regimes where plain INT2 remains relatively stable.

\subsection{Prompt-length and chunk-size breakdown}
\label{app:llm_bins}

\begin{figure}[t]
  \centering
  \includegraphics[width=\textwidth]{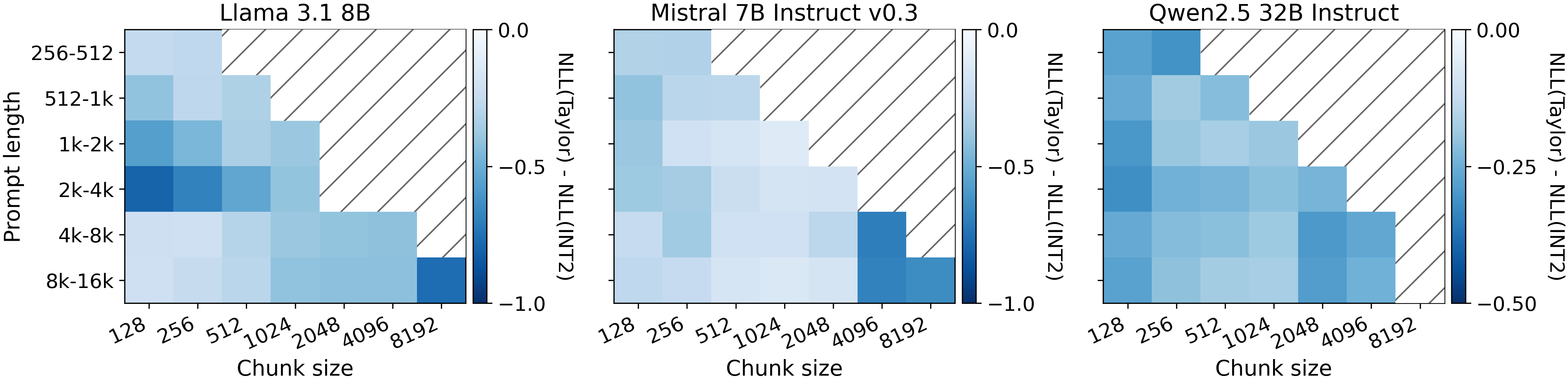}
  \caption{
  Prompt-length and chunk-size breakdown for LLM partial-prefill experiments. The plotted value is $\mathrm{NLL}_{\mathrm{INT2+Taylor}} - \mathrm{NLL}_{\mathrm{INT2}}$, computed within matched model, chunk-size, prompt-bin, and evaluation-example cells. Negative values indicate that the Taylor correction reduces teacher-forced NLL relative to plain INT2 KV-cache quantization. Striped areas indicate no available matched data.
  }
  \label{fig:llm_paired_nll_heatmap}
\end{figure}

To test whether the aggregate results are driven by a small subset of prompt lengths, we also analyze NLL by retained prompt-length bin. Figure~\ref{fig:llm_paired_nll_heatmap} reports paired NLL differences grouped by prompt-length bin and current-chunk size.

\subsection{Attention-mass diagnostic}
\label{app:llm_attention_mass}

\begin{figure}[t]
  \centering
  \includegraphics[width=\textwidth]{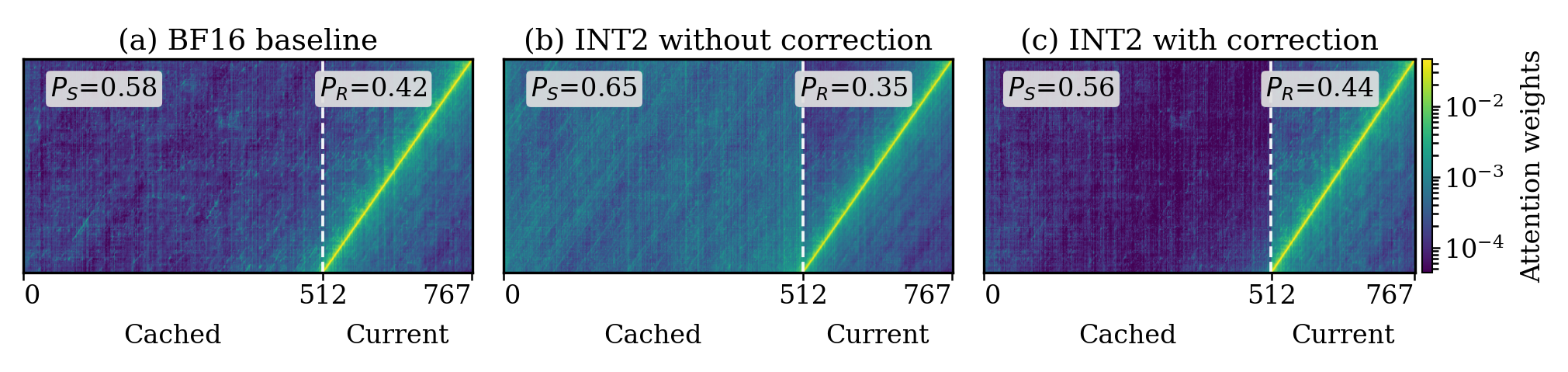}
  \caption{
  Attention weights for Llama-3.2-1B under INT2 KV-cache quantization. The visualized attention weights are averaged over representative prompts with lengths in $[1024, 2048)$, layers, and attention heads for chunk size $256$. The dashed vertical line separates cached-prefix tokens from current-chunk tokens. Panel \textbf{(b)} shows that, relative to the BF16 baseline in \textbf{(a)}, quantization increases attention weights in the cached block of tokens and decreases them in the current chunk. This effect is quantified by the attention masses $P_\mathcal{S}$ and $P_\mathcal{R}$ of the cached token block and current chunk. Panel \textbf{(c)} shows that our correction largely restores the original attention weights, with slight overcorrection.
  }
  \label{fig:llm_motivation_distribution}
\end{figure}

The central mechanism studied in the main paper is that quantized cached keys receive inflated softmax mass because the exponential transforms zero-mean score noise into a positive partition-sum bias (Fig.~\ref{fig:jensen_bias}; see also Fig.~\ref{fig:motivation_distribution}). Figure~\ref{fig:llm_motivation_distribution} visualizes the corresponding attention-weight shift in an LLM partial-prefill setting.

For this diagnostic, we use Llama-3.2-1B as a lightweight model for attention visualization. This diagnostic model is separate from the three-model NLL benchmark above; it is used here because logging full attention weights across many layers, heads, prompts, and chunk sizes is memory intensive.

\subsection{Discussion}
\label{app:llm_discussion}

The LLM partial-prefill results provide additional indication in a cached/current attention structure setting similar to the main experiments on chunked auto-regressive video diffusion in \ref{sec:experiments}. In the completed paired comparisons, plain INT2 KV-cache quantization generally worsens teacher-forced NLL, while the Taylor correction reduces NLL relative to plain INT2. This trend is consistent with our derivation and video-model experiments, but we interpret the LLM results as a diagnostic extension rather than as a comprehensive LLM KV-cache quantization benchmark. We therefore emphasize paired teacher-forced NLL comparisons and leave optimized LLM kernels, broader task-level evaluation, and attention-mass diagnostics across more LLM models and chunk sizes to future work.

In some configurations, the corrected condition obtains lower NLL than the BF16 baseline. We treat this observation cautiously and do not interpret it as a general improvement over BF16. It may depend on the partial-prefill setup, the teacher-forced NLL objective, or mild overcorrection from the Taylor approximation at aggressive bitwidths. Our main conclusion from these experiments is limited to the paired comparison between plain INT2 and INT2 with correction: the correction reduces the NLL degradation introduced by INT2 KV-cache quantization in the evaluated partial-prefill settings.

\section{Broader Impact}

This work proposes a training-free correction for KV-cache quantization in autoregressive video diffusion models. The direct goal is to improve the efficiency and quality of long-form video generation by reducing memory usage while preserving generation fidelity. Potential positive impacts include lowering the computational cost of research on long-video and world-model generation, enabling longer context windows under fixed memory budgets, and improving accessibility of efficient inference methods for academic and resource-constrained settings.

At the same time, improvements in the efficiency and fidelity of video generation may also lower the cost of generating synthetic video content. As with other advances in generative video modeling, this could indirectly facilitate misuse such as producing misleading synthetic media, impersonation, or disinformation. Our work does not introduce a new generative model, dataset, or training procedure, and we do not release new model weights. The method is an inference-time numerical correction applied to existing models, so the primary risks are inherited from the underlying video generation systems on which it is used. We encourage deployment only in settings that follow the safety policies, watermarking or provenance mechanisms, and misuse-monitoring practices appropriate for the underlying generative model.

% \clearpage
% \input{checklist.tex}

\end{document}